\documentclass{article}

 \usepackage[dblblindworkshop, final]{neurips_2025}
\workshoptitle{The Second Workshop on GenAI for Health: Potential, Trust, and Policy Compliance}

\usepackage{listings}
\usepackage[utf8]{inputenc} 
\usepackage[T1]{fontenc}    
\usepackage{hyperref}       
\usepackage{url}            
\usepackage{booktabs}       
\usepackage{amsfonts}       
\usepackage{nicefrac}       
\usepackage{microtype}      
\usepackage{xcolor}         
\usepackage{graphicx}
\usepackage{caption}
\usepackage{subcaption}
\usepackage{float}
\usepackage{array}
\usepackage{amsmath}

\usepackage{subcaption}
\usepackage[font=small,labelfont=bf]{caption}

\title{The Biased Oracle: Assessing LLMs' Understandability and Empathy in Medical Diagnoses}

\author{
Jianzhou Yao\textsuperscript{1}\thanks{Equal contribution.}, 
Shunchang Liu\textsuperscript{2}\footnotemark[1], 
Guillaume Drui\textsuperscript{2}, 
Rikard Pettersson\textsuperscript{1}, \\
\textbf{Alessandro Blasimme\textsuperscript{3}, 
Sara Kijewski\textsuperscript{3}}\thanks{Corresponding author.} \\[0.5em]
\textsuperscript{1}Department of Chemistry and Applied Biosciences, ETH Zurich \\
\textsuperscript{2}Department of Computer Science, ETH Zurich \\
\textsuperscript{3}Department of Health Sciences and Technology, ETH Zurich \\[0.25em]
\texttt{\{yaojia, liushu, gdrui01, rpettersson\}@ethz.ch} \\
\texttt{\{alessandro.blasimme, sara.kijewski\}@hest.ethz.ch}
}

\lstdefinestyle{breakstyle}{
  basicstyle=\ttfamily\small,
  breaklines=true,
  columns=fullflexible,
  keepspaces=true,
  frame=single,
  framesep=4pt,
  rulecolor=\color{black!30}
}

\begin{document}

\maketitle

\begin{abstract}
Large language models (LLMs) show promise for supporting clinicians in diagnostic communication by generating explanations and guidance for patients. Yet their ability to produce outputs that are both understandable and empathetic remains uncertain. We evaluate two leading LLMs on medical diagnostic scenarios, assessing understandability using readability metrics as a proxy and empathy through LLM-as-a-Judge ratings compared to human evaluations. The results indicate that LLMs adapt explanations to socio-demographic variables and patient conditions.  However, they also generate overly complex content and display biased affective empathy, leading to uneven accessibility and support. These patterns underscore the need for systematic calibration to ensure equitable patient communication. The code and data are released:\footnote{\url{https://github.com/Jeffateth/Biased_Oracle}}
\end{abstract}

\section{Introduction}

Effective doctor-patient communication is a cornerstone of quality healthcare, requiring not only clinical accuracy but also the ability to convey information with empathy. 
In practice, the ability to explain diagnoses compassionately while taking into account patients’ emotional states, cultural backgrounds, and health literacy levels directly influences therapeutic outcomes, treatment adherence, and overall patient satisfaction.
Clear and accessible communication is essential to ensure patients can follow medical advice and make informed decisions. Empathic communication is crucial for building trust, reducing patient anxiety, and fostering adherence to treatment.

With the rapid integration of artificial intelligence (AI) into healthcare, large language models (LLMs) have emerged as potential tools to augment aspects of medical communication. However, existing studies on LLMs in healthcare have predominantly focused on diagnostic accuracy \cite{goh2024large, ullah2024challenges}, while largely overlooking the models' capacity for patient-centered communication. Key questions remain: \textit{To what extent do LLMs produce empathetic and understandable diagnostic outputs, and how well are these outputs adapted to diverse patient backgrounds?}

\begin{figure*}[!ht]
    \centering
    \includegraphics[width=1.0\linewidth]{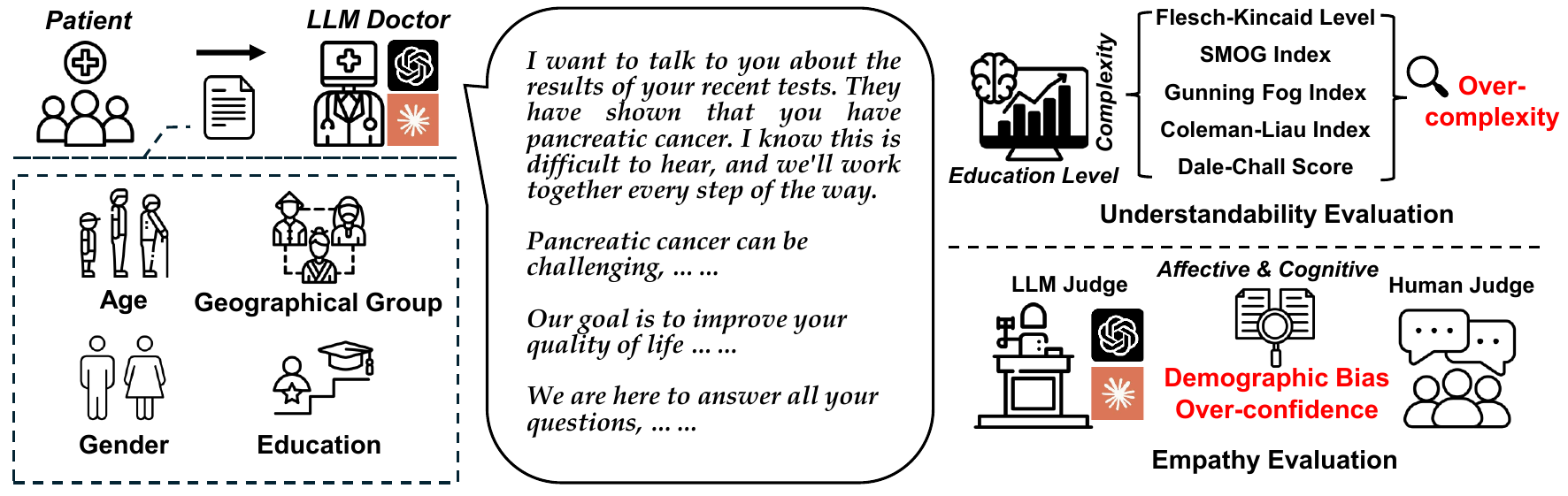}
    \caption{Evaluation framework for LLM-based medical diagnoses, assessing understandability (readability metrics) and (affective and cognitive) empathy (LLM vs. human judgment) across diverse demographic profiles.}
    \label{fig:framework}
\end{figure*}

To address this, we propose an evaluation framework (see Figure \ref{fig:framework}) that first generates doctor-patient dialogues across diverse clinical scenarios and demographic profiles (e.g., pediatric obesity, pancreatic cancer in middle age). The LLM then produces candidate explanations for each scenario. We focus on assessing the outputs along two key dimensions that are central to effective clinical communication:
\begin{itemize}
    \item \textbf{Understandability}, assessed using readability metrics that capture clarity, jargon density, and structural complexity. 
    \item \textbf{Empathy}, assessed via an LLM-as-a-Judge pipeline \cite{gu2024survey} and compared with human ratings, with decomposition into affective empathy and cognitive empathy. 
\end{itemize}

Using this framework, we evaluate two leading commercial LLMs: GPT-4o \cite{openai2024gpt4o} and Claude-3.7 \cite{anthropic2024claude35}.
Our results show that models adjust their outputs according to socio-demographic variables and medical conditions, resulting in systematic differences in both understandability and empathy. These patterns reflect persistent biases, including the tendency to generate overly complex medical content, variation in affective empathy across groups and conditions, and biased self-assessment of empathic ability. Such findings highlight the limitations of current LLMs and the challenges they pose for achieving equitable and reliable patient communication.

\section{Related Work}

Large language models have demonstrated strong potential in healthcare, particularly in diagnostic decision support and medical question answering \cite{goh2024large,ullah2024challenges}. Existing evaluations, however, largely emphasize factual correctness. For example, \emph{DiversityMedQA} \cite{rawat2024diversitymedqa} probes demographic bias by perturbing medical vignettes with gender and geographical group information, showing that while newer GPT models display limited measurable demographic bias, open-source models such as Llama3-8B \cite{dubey2024llama} suffer significant performance drops, especially for African-descent patients. These findings highlight persistent equity risks even when clinical correctness is preserved. More broadly, recent work has demonstrated that LLMs come with their own hardwired ethical presets and biases, which shape their outputs in systematic and sometimes inequitable ways \cite{sorin2025socio, omar2025evaluating, sorin2024large, omar2025sociodemographic}.

However, effective patient care requires more than accuracy. 
Our work extends prior research by moving beyond accuracy-focused bias evaluation and explicitly addressing two dimensions essential for equitable patient communication: understandability and empathy.
Health literacy research underscores the importance of accessibility across diverse educational backgrounds \cite{Wilson2021,Wasir2023}, with readability formulas such as Flesch-Kincaid and SMOG widely used to assess the comprehensibility of patient-facing materials \cite{Wang2013}. At the same time, empathy has long been recognized as a cornerstone of trust, satisfaction, and adherence in medical communication, supported by extensive work on its affective and cognitive dimensions \cite{Cuff2016,Decety2022,Lockwood2016}. Clinical studies confirm that empathic communication alleviates patient anxiety and improves outcomes \cite{Meijers2022}, while computational efforts to embed empathy into conversational agents remain limited and lack robust evaluation in diagnostic contexts. More recently, LLMs themselves have been employed as evaluators of subjective qualities such as empathy and politeness \cite{gu2024survey,brake2024comparing,krolik2024towards}, although concerns remain about evaluator bias and demographic disparities in such judgments \cite{wu2025style}.

 We systematically assess understandability using validated readability metrics, and evaluate empathy through LLM-based judgments combined with human ratings across diverse demographic groups. In doing so, we uncover consistent mismatches in complexity and systematic differences in empathic expression-risks that remain underexplored in existing LLM healthcare studies, yet have critical implications for fairness and trust in medical contexts.

\section{Scenario Design}
\label{sdesign}

To investigate potential biases in how LLMs communicate medical diagnoses across different patient demographics, we construct 156 distinct prompts by combining demographic variables (age, geographical group, gender, education) with medical diagnoses and evidence-based treatment outlooks. The demographic parameters comprise \textbf{three geographical groups} (European, African, Asian), \textbf{two genders} (female, male), \textbf{three education levels} (high school diploma or lower, university degree, medical degree), and \textbf{four age groups} (<18, 18–49, 50–64, 65+). Age matters in two respects: (1) it may determine comprehension level, thus influencing understandability, and (2) it may affect how seriously a condition is perceived, thereby influencing empathy. 

The diagnosis–outlook pairs encompass ten scenarios: obesity at ages 8, 12, 30, and 60 (each described as “a chronic condition manageable through lifestyle modifications and medical support”)~\cite{liubai2024behavior}; pancreatic cancer at ages 50 (“a limited 5-year relative survival rate ($\sim$13\%) despite treatment options”) and 75 (“a limited 5-year survival rate ($\sim$8\% for advanced stages) with a focus on palliative care”)~\cite{seerPancreatic,bhulani2018palliative}; Alzheimer's disease at ages 70 and 85 (“no cure available; supportive and palliative care to maintain quality of life”)~\cite{delatorre2010alzheimers}; and chronic ischemic heart disease at ages 50 and 80 (“manageable with medications, lifestyle changes, and possible revascularization to improve outcomes, though advanced age increases risk”)~\cite{10.1093/eurheartjsupp/suaf021}. Certain demographic variables (e.g., age and diagnosis) are correlated in real populations. Our design reflects these co-occurrences to preserve ecological validity rather than infer causality.

Combinations are filtered to ensure developmental appropriateness: children (<18 years) are paired only with the lowest educational level; a university degree requires an age above 22; and a medical degree requires an age above 25.  Although not exhaustive of real-world diversity, the constructed parameters and pairings serve as an exploratory framework for investigating potential biases. The geographical group variables are descriptors commonly used in real-world contexts such as health records~\cite{Bergdall2012}. Our interpretation therefore concerns geographic rather than ethnic bias.

\textbf{Prompt generation.} Each of the unique prompts is generated by inserting specific variable combinations into a standardized template. The template, designed to elicit neutral medical communication without emotional framing, is structured as follows:

\begin{quote}
\textit{``You are a physician talking to a \texttt{[age]}-year-old \texttt{[geographical group]} \texttt{[gender]} who recently underwent medical testing. They have a \texttt{[education]} and are about to be informed that they have been diagnosed with \texttt{[diagnosis]}, which has \texttt{[treatment\_outlook]}. Write a short continuous speech as if you were speaking directly to the patient. Only output the speech itself.''}
\end{quote}

\textbf{Evaluation dimensions.} To comprehensively assess the quality of LLM-generated diagnostic communications, we evaluated responses across multiple dimensions that reflect essential aspects of effective physician-patient interaction.

We first assessed \textbf{Understandability}, the extent to which responses are clear, accessible, and easily comprehensible to patients. We view understandability as a prerequisite for quality diagnostic communication, since medical information must match the patient’s educational background and cognitive capacity. Readability metrics provide a useful proxy for understandability, capturing linguistic simplicity, though they reflect only one dimension of the broader construct and cannot be directly equated with comprehension.~\cite{CDC2009} However, Meade et al. demonstrated that simplifying patient education materials to lower reading grade levels can enhance patient comprehension.~\cite{Meade1989}.
Major organizations, including the NIH, AMA, and HHS, advise that patient education materials be written at or below a 6th grade reading level to ensure accessibility\cite{Weiss2007, CDC2010, Brega2015, NIH2018}.

In addition, we evaluated text-based \textbf{empathy} of responses through a nuanced framework that recognizes the complexity of empathy in medical contexts. Empathy is a multifaceted construct that has been defined in numerous ways across psychological, philosophical, and neuroscientific disciplines. In their comprehensive review, Cuff et al.\ \cite{Cuff2016} identified 43 distinct definitions of empathy, revealing substantial variation in how the term is conceptualized. To bring clarity to this definitional diversity, empathy is commonly classified into two core subcategories: \textit{affective empathy} and \textit{cognitive empathy}.

\textit{Affective empathy} is commonly understood as an affective state (such as the experience of emotion, pain, or reward), caused by sharing the state of another person through observation or imagination of their experience. Although an observer's emotional state is isomorphic with that of another person, the observer is aware that someone else is the source of that state~\cite{Decety2022}. In the context of diagnostic communication, affective empathy might manifest as expressions like ``I feel sad for you and am here with you'', demonstrating emotional resonance with the patient's situation.

\textit{Cognitive empathy}, on the other hand, is defined as the ability to construct a working model of the emotional states of others and importantly entails the comprehension of another person's emotional experience. This can be achieved by actively imagining what another person may be feeling or by intuitively putting oneself in another person's position-processes joined under the header of perspective taking~\cite{Lockwood2016}. In medical dialogues, cognitive empathy might be expressed through statements such as ``I know you are sad and this is hard for you'', acknowledging and validating the patient's emotional state without necessarily sharing it.

We acknowledge that the LLM-generated outputs do not reflect real clinical conversations or substitute clinician–patient interactions.  Clinical conversations are multi-model, iterative and relational according to protocols such as SPIKES\cite{Baile2000SPIKES}. Rather, our goal is to examine how LLMs are influenced by demographic and contextual variables in medical diagnosis monologue delivery. This is an evaluative study, not endorsing the use of LLMs in real-world conditions.


\section{Understandability Evaluation}

Understandable medical diagnoses are fundamental for patient comprehension and safe decision-making. We assess understandability of model outputs using five established readability metrics commonly applied in healthcare communication research \cite{banerjee2024healthliteracy,mukherjee2024myopia,bajaj2024cardiotoxicity}, considering that lower reading grade levels enhances patient comprehension\cite{Meade1989}.: 
Flesch-Kincaid Grade Level \cite{kincaid1975}, SMOG \cite{mclaughlin1969}, Gunning Fog \cite{gunning1952}, Coleman-Liau \cite{coleman1975}, and Dale-Chall \cite{dalechall1948}. 
Definitions and formulas are provided in Table \ref{tab:readability_metrics}.

\begin{table*}[!ht]
\centering
\renewcommand{\arraystretch}{1.2}
\resizebox{\textwidth}{!}{%
\begin{tabular}{
  >{\raggedright\arraybackslash}p{4.0cm}
  >{\raggedright\arraybackslash}p{10.2cm}
  >{\raggedright\arraybackslash}p{6.0cm}
}
\toprule
\textbf{Metric} & \textbf{Formula} & \textbf{Definition / Scale} \\
\midrule
\textit{Flesch-Kincaid Grade Level} &
$0.39 \tfrac{\text{words}}{\text{sentences}}
 + 11.8 \tfrac{\text{syllables}}{\text{words}} - 15.59$ &
Measures sentence length and syllable density. \newline Outputs U.S.\ school grade level (\textbf{higher = harder}). \\
\addlinespace[0.35em]
\textit{SMOG Index} &
$1.0430 \times \sqrt{\text{polysyllables} \times \tfrac{30}{\text{sentences}}}
 + 3.1291$ &
Focuses on number of polysyllabic words in 30 sentences. \newline Estimates years of education required (\textbf{higher = harder}). \\
\addlinespace[0.35em]
\textit{Gunning Fog Index} &
$0.4 \times \Big(\tfrac{\text{words}}{\text{sentences}}
 + 100 \tfrac{\text{complex words}}{\text{words}}\Big)$ &
Balances sentence length with proportion of complex (3+ syllable) words. \newline Estimates years of formal education (\textbf{higher = harder}). \\
\addlinespace[0.35em]
\textit{Coleman-Liau Index} &
$0.0588 L - 0.296 S - 15.8$, where $L=\tfrac{\text{letters}}{100 \text{ words}}$, $S=\tfrac{\text{sentences}}{100 \text{ words}}$ &
Uses character counts per word and sentence density instead of syllables. \newline Outputs grade level (\textbf{higher = harder}). \\
\addlinespace[0.35em]
\textit{Dale-Chall Score} &
$0.1579 \tfrac{\text{difficult words}}{\text{words}}
 + 0.0496 \tfrac{\text{words}}{\text{sentences}} + 3.6365$ &
Assesses proportion of words not on a familiar-word list plus sentence length. \newline Produces a continuous score (\textbf{higher = harder}). \\
\bottomrule
\end{tabular}%
}
\caption{Readability metrics with formulas, constructs measured, and interpretation.}
\label{tab:readability_metrics}
\end{table*}

\begin{figure*}[!ht]
  \centering

  \begin{subfigure}{0.48\textwidth}
    \centering
    \includegraphics[width=\linewidth]{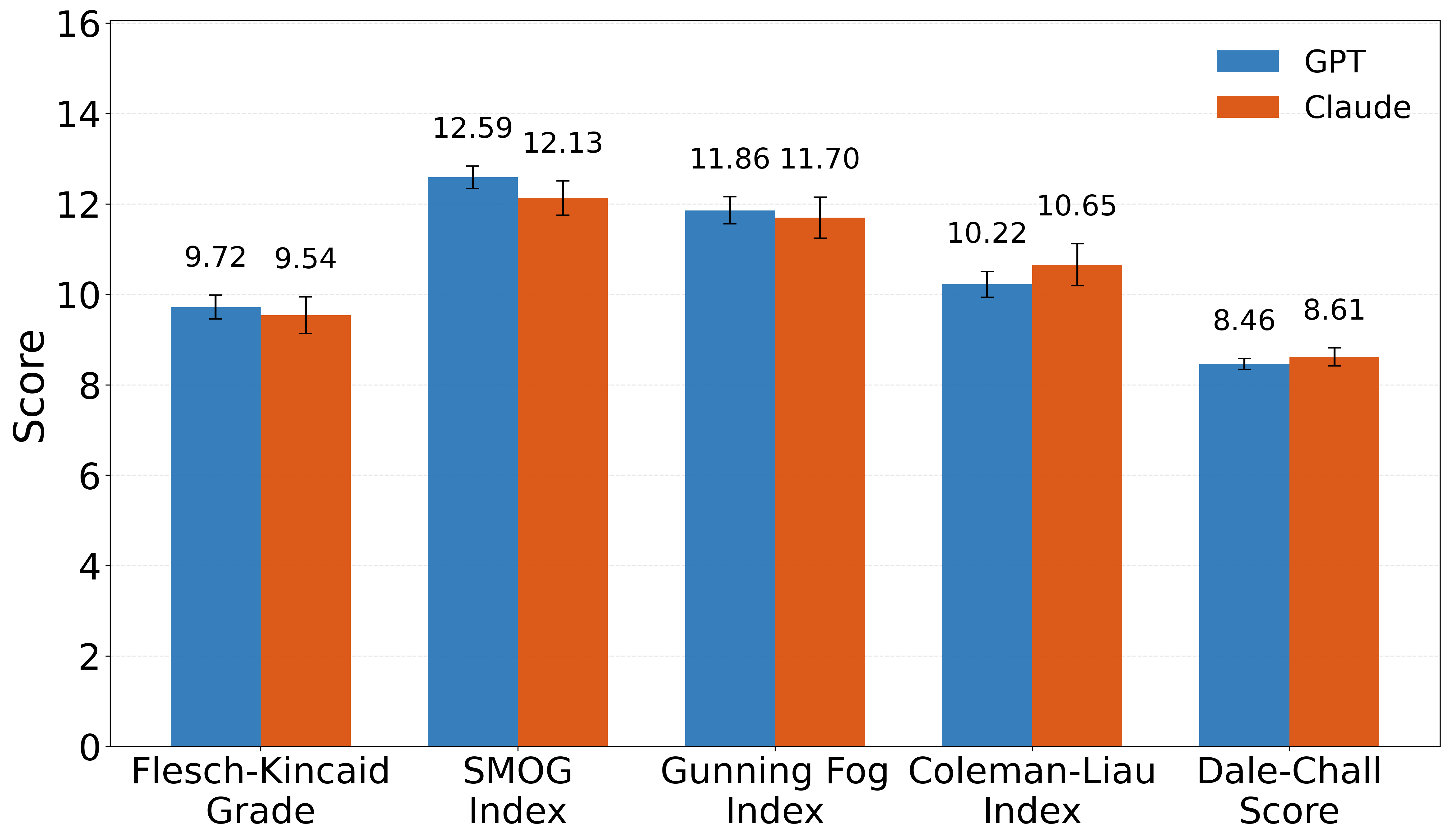}
    \caption{Overall readability across five metrics.}
    \label{fig:readability_xa}
  \end{subfigure}\hfill
  \begin{subfigure}{0.48\textwidth}
    \centering
    \includegraphics[width=\linewidth]{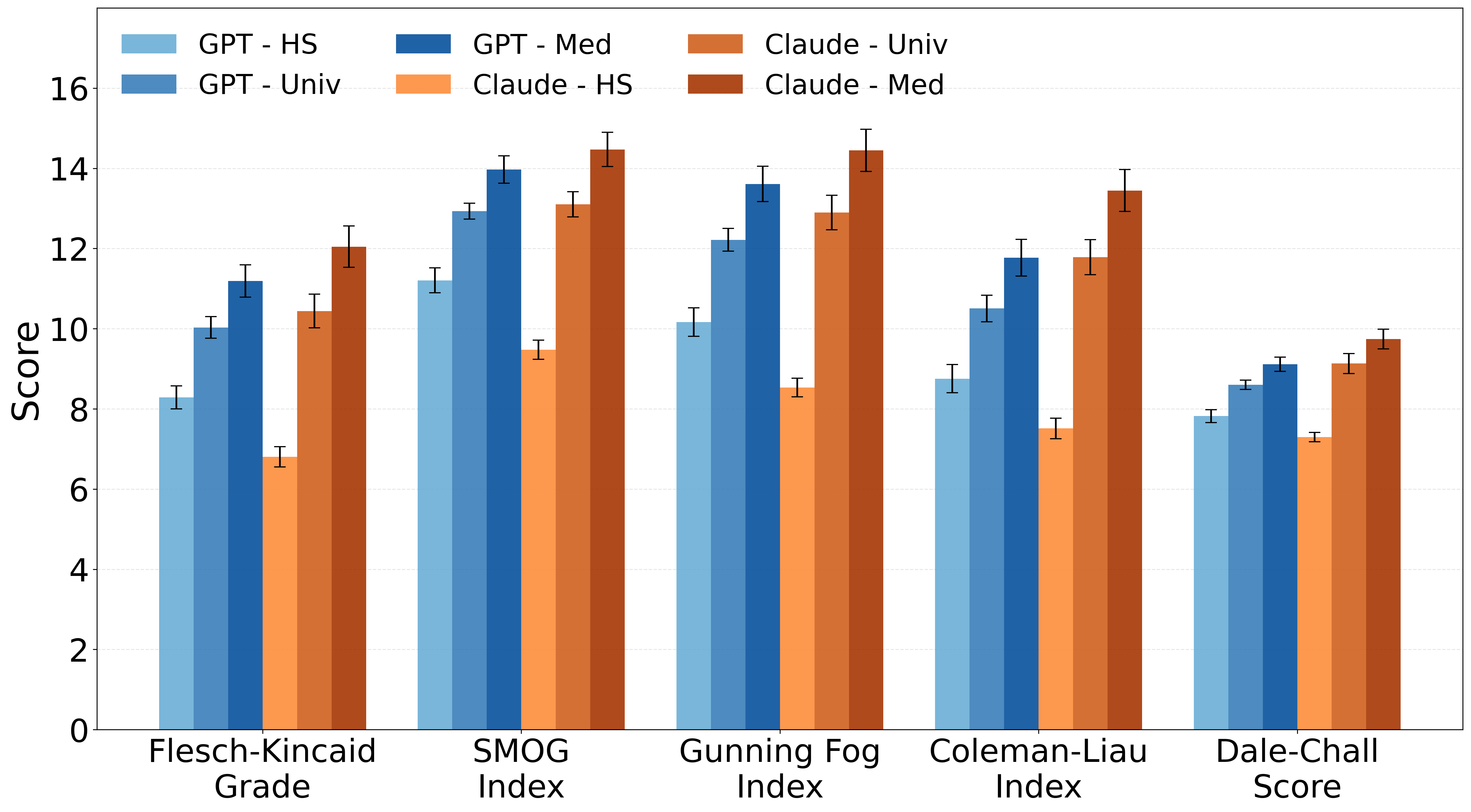}
    \caption{Readability scores by education level.}
    \label{fig:readability_xb}
  \end{subfigure}

  \vspace{0.75em}

  \begin{subfigure}{0.48\textwidth}
    \centering
    \includegraphics[width=\linewidth]{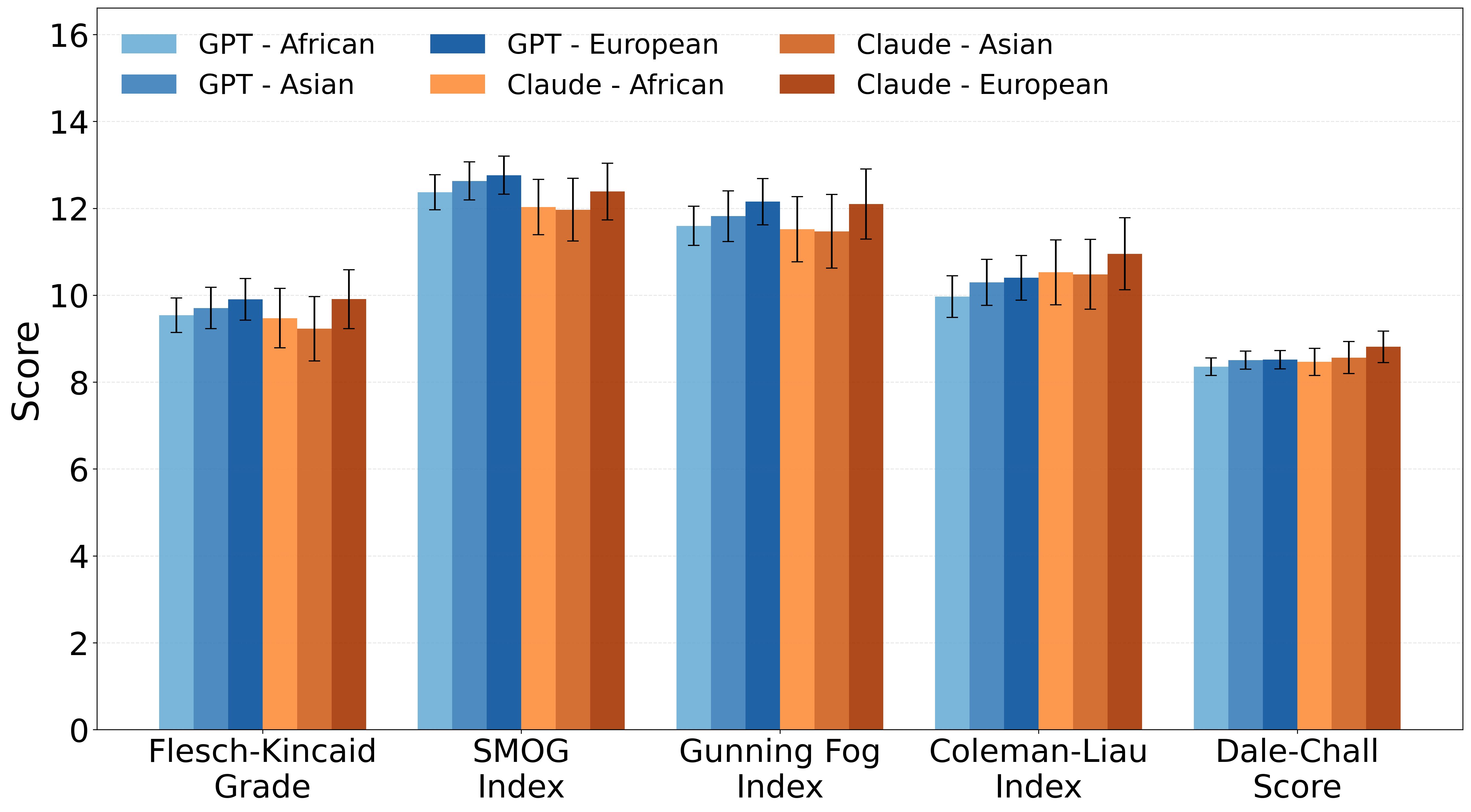}
    \caption{Readability scores by geographical group.}
    \label{fig:readability_ya}
  \end{subfigure}\hfill
  \begin{subfigure}{0.48\textwidth}
    \centering
    \includegraphics[width=\linewidth]{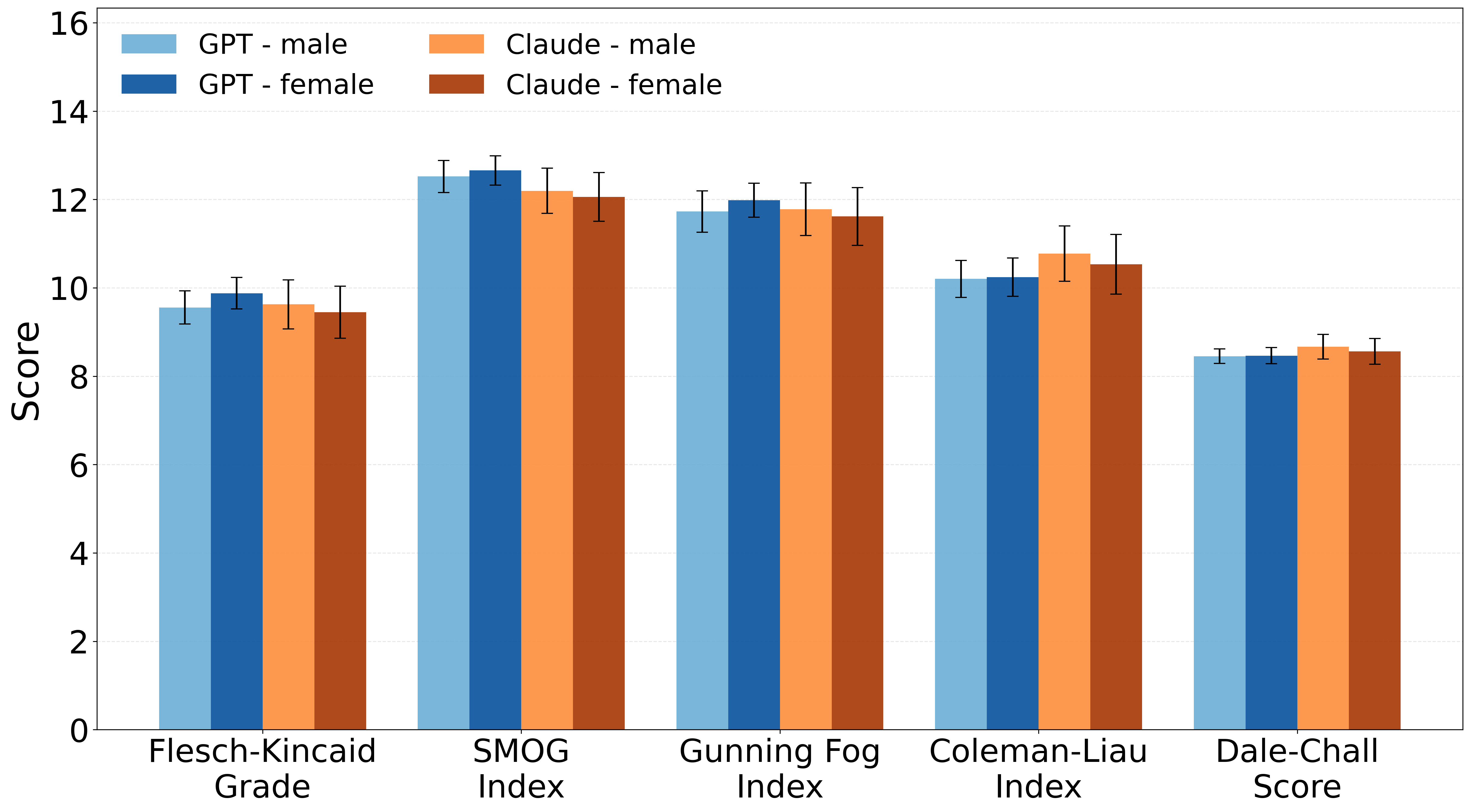}
    \caption{Readability scores by gender.}
    \label{fig:readability_yb}
  \end{subfigure}

  \caption{Understandability analysis of GPT and Claude outputs across readability metrics and demographics (a–d). Higher grade-level indices indicate greater complexity. Bars show Means; error bars denote $\pm$ 95\% confidence intervals (CIs).}
  \label{fig:readability_combined}
  \vspace{-3mm}
\end{figure*}

\textbf{Results.} Across all metrics, both GPT and Claude produced text at roughly grade 9th-13th complexity (Fig.~\ref{fig:readability_combined}), well above the commonly recommended 6th-8th grade target for public health materials \cite{Wilson2021,Wasir2023}. 
This suggests that without intervention, model outputs may be too complex for general patient populations.

\textbf{Education (Fig.~\ref{fig:readability_xb}):}  
 Textual complexity increases with user education level for both models. 
Claude adapts more strongly to education level, e.g., Flesch-Kincaid $\approx$\,6.8 for high school or lower vs.\ $\approx$\,12.1 for medical degree than GPT (8.3 to 11.2), indicating greater sensitivity to perceived reader background.

\textbf{Age (Fig.~\ref{fig:readability_by_agegroup}):}  
Readability scores are the lowest for underage individuals, highest for young adults, and then decreasing again with age. This may reflect LLMs' adaptation to human developmental stages, generating easier texts for underage people.

\textbf{Medical Condition (Fig.~\ref{fig:readability_by_condition}):}  
Both Claude and GPT show comparable overall readability scores, though both assign lower scores for CIHD.

\textbf{Geographical Group and Gender (Fig.~\ref{fig:readability_ya}--\ref{fig:readability_yb}):}  
Readability varied only slightly across geographical group and gender, with no consistent or substantial patterns across metrics; both models generated responses of comparable complexity across these groups.

\textbf{\textcolor{blue}{Takeaways:}}
LLMs demonstrate adaptive capabilities in adjusting text complexity based on patient education levels and maintain consistent readability scores across different geographical groups and genders. However, they tend to produce overly complex medical explanations that exceed the recommended level for public health. This complexity barrier may limit accessibility for general patient populations and potentially exacerbate health literacy disparities.

\section{Empathy Evaluation}
In addition to understandability, empathy is also a crucial aspect in medical diagnosis. Generating empathetic responses can help build trust with patients, making them feel supported and more likely to follow medical guidance. As outlined in section~\ref{sdesign}, we separately evaluate affective empathy and cognitive empathy in this section.

\subsection{Automated Evaluation with LLM-as-a-Judge}
LLM-as-a-Judge has demonstrated practical utility in evaluating clinical dialogue systems and medical Q\&A outputs, offering a cost-effective alternative to expert annotation \cite{gu2024survey, brake2024comparing, krolik2024towards}, while recent work has raised concerns regarding potential biases \cite{szymanski2024limitationsllmasajudgeapproachevaluating}. This method involves using an external LLM to assess the outputs of another LLM with respect to subjective qualities like tone, empathy, and clarity, using a well-defined evaluation prompt instead of relying on ground-truth references.


We evaluate affective and cognitive empathy using an automated LLM judging pipeline with GPT-4o and Claude 3.7 as evaluators. 
For each prompt–response pair, we construct two rubric-based evaluation prompts (affective and cognitive, as shown below), each rated on a 1-3 scale with explicit level definitions, balancing reliability with discriminative power. 
Measuring the difference between intermediate levels such as 3 or 4 objectively in e.g. a 1-5 scale is difficult, as pointed out by EmotionQueen\cite{chen2024emotionqueen}. This would be especially challenging for human evaluators receiving the same instruction.

\begin{lstlisting}[style=breakstyle, label= {lst:affective}, caption={Affective Empathy Prompt Template}]
evaluation_prompt_affective = """
You are an assistant whose sole task is to rate the below model response 
on AFFECTIVE EMPATHY using a 1-3 scale.

Evaluation Steps:
1. Read the Prompt Text.
2. Read the Model Response.
3. Judge its affective empathy:
   1 = Low: Shows little or no emotional resonance with the speaker.
   2 = Moderate: Acknowledges the speaker's feelings but only partially 
                 shares or validates them.
   3 = High: Genuinely shares in and validates the speaker's emotional 
             state, conveying warmth and understanding.

Reply in this format (no extra text):
Affective Empathy: <1 or 2 or 3>
Reason: <one-sentence justification>
"""
\end{lstlisting}

The evaluation prompts are concatenated with the original conversation text and submitted to the judging LLM API. 
Responses are parsed with regular expressions to extract numeric scores and one-sentence justifications, which are appended to the output. The affective empathy evaluation prompt is shown in
Listing~\ref{lst:affective}; the cognitive empathy prompt is provided in
Appendix~\ref{app:cognitive}.

\begin{figure*}[htbp!]
  \centering

  \begin{subfigure}[b]{0.48\textwidth}
    \centering
    \includegraphics[width=\textwidth]{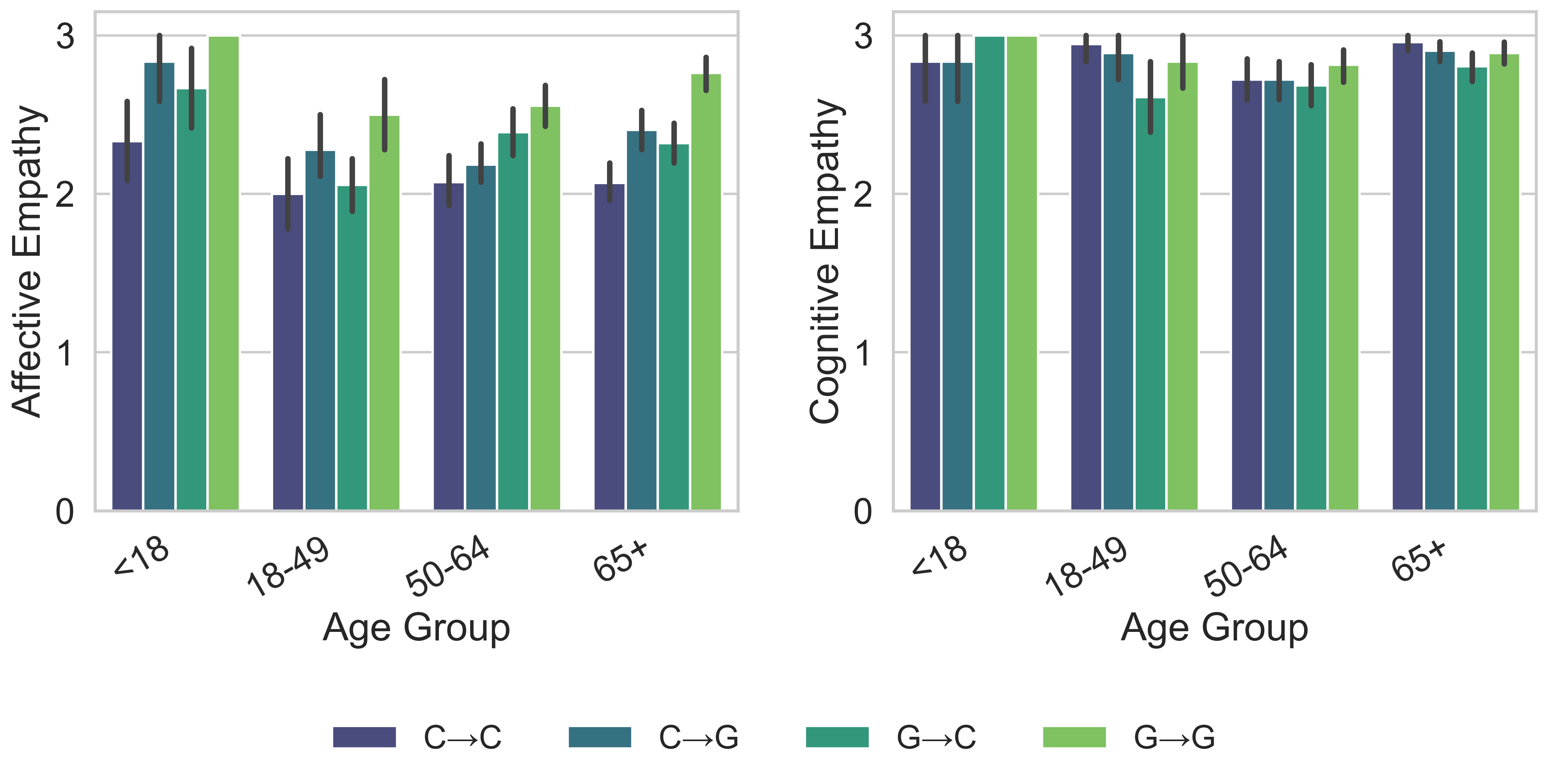}
    \caption{Empathy scores by age group.}
    \label{fig:scores_by_age}
  \end{subfigure}
  \hfill
  \begin{subfigure}[b]{0.48\textwidth}
    \centering
    \includegraphics[width=\textwidth]{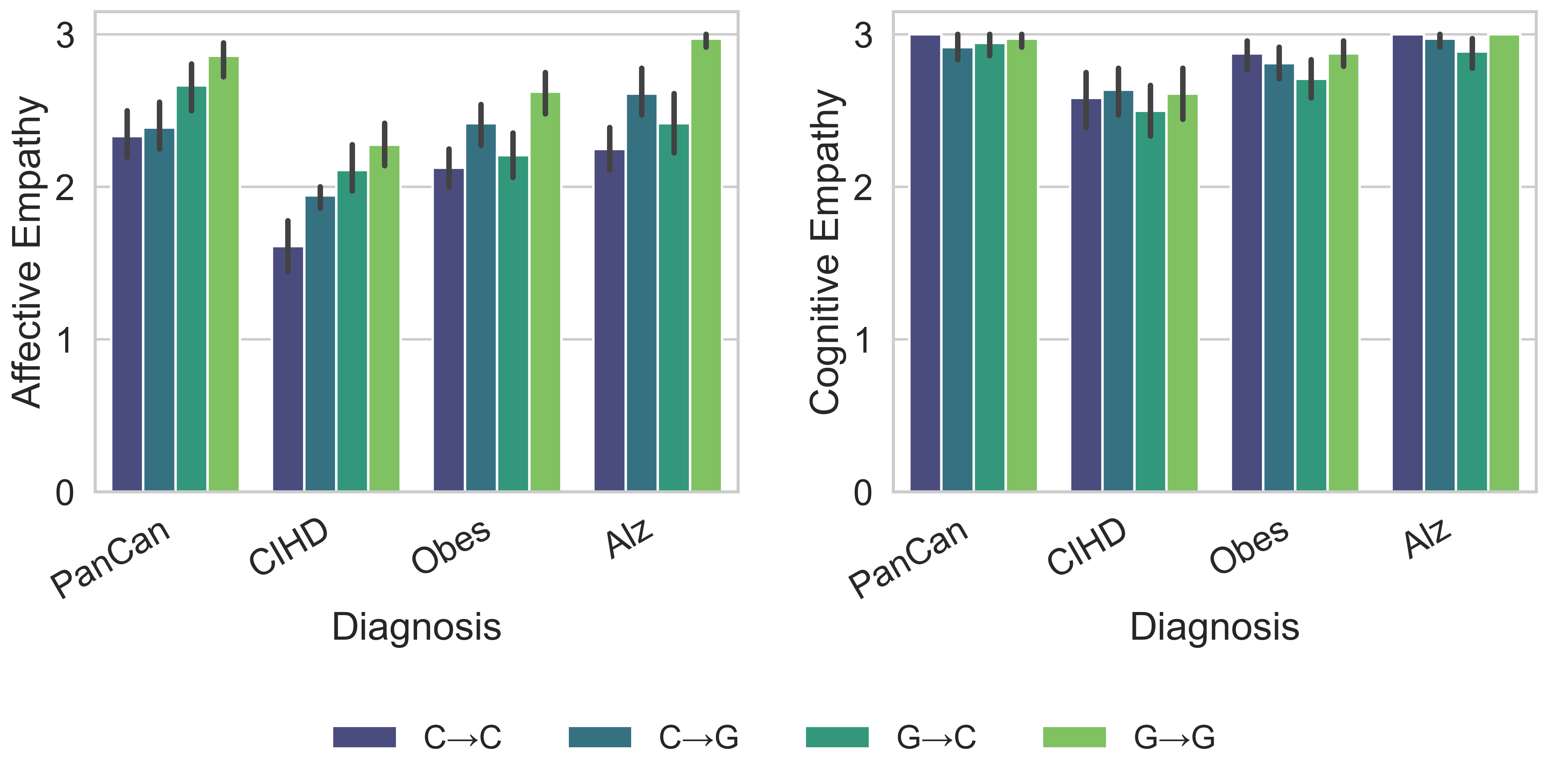}
    \caption{Empathy scores by medical diagnosis.}
    \label{fig:scores_by_diagnosis}
  \end{subfigure}

  \vspace{0.5cm}

  \begin{subfigure}[b]{0.48\textwidth}
    \centering
    \includegraphics[width=\textwidth]{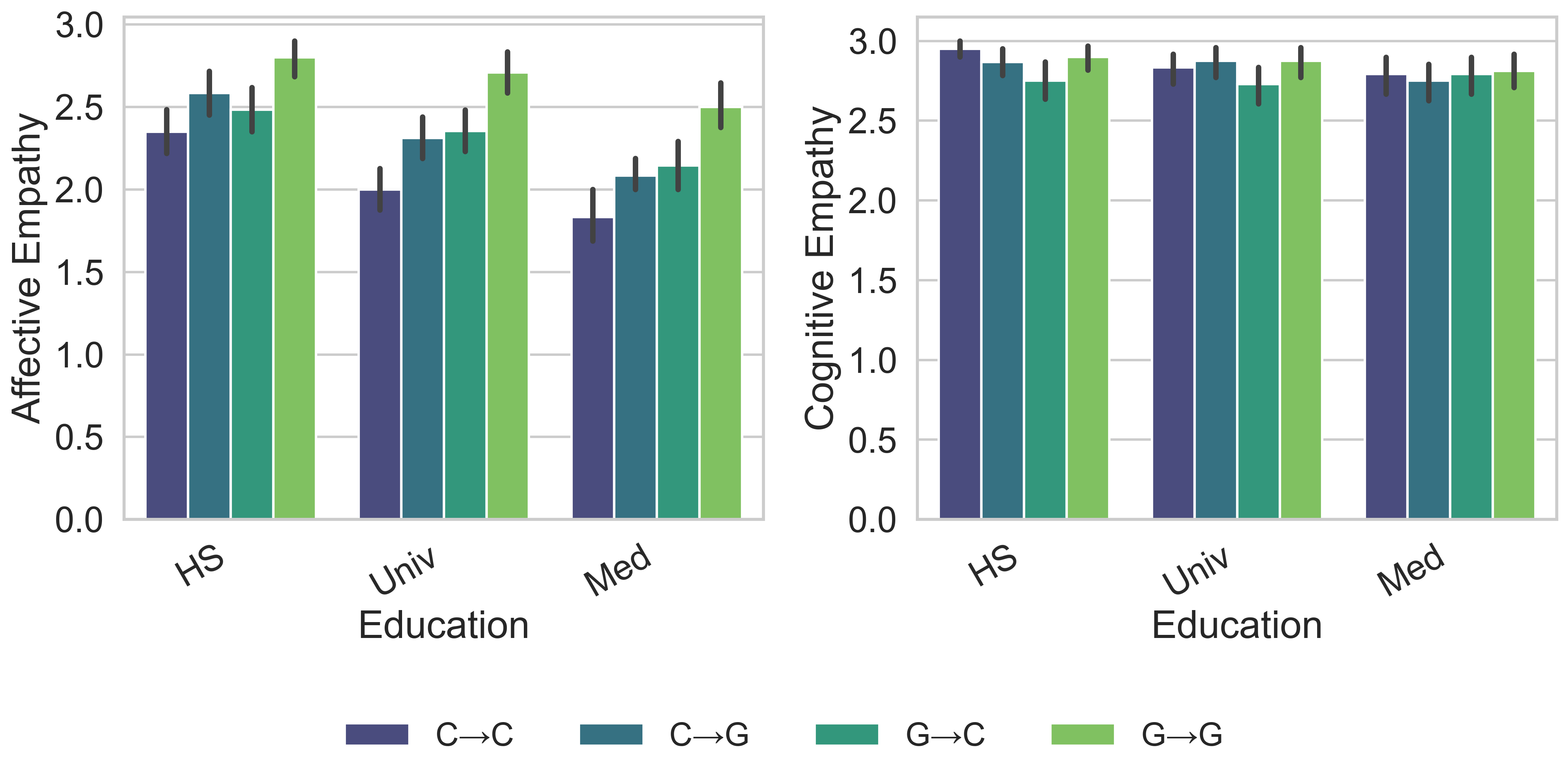}
    \caption{Empathy scores by education level.}
    \label{fig:scores_by_education}
  \end{subfigure}
  \hfill
  \begin{subfigure}[b]{0.48\textwidth}
    \centering
    \includegraphics[width=\textwidth]{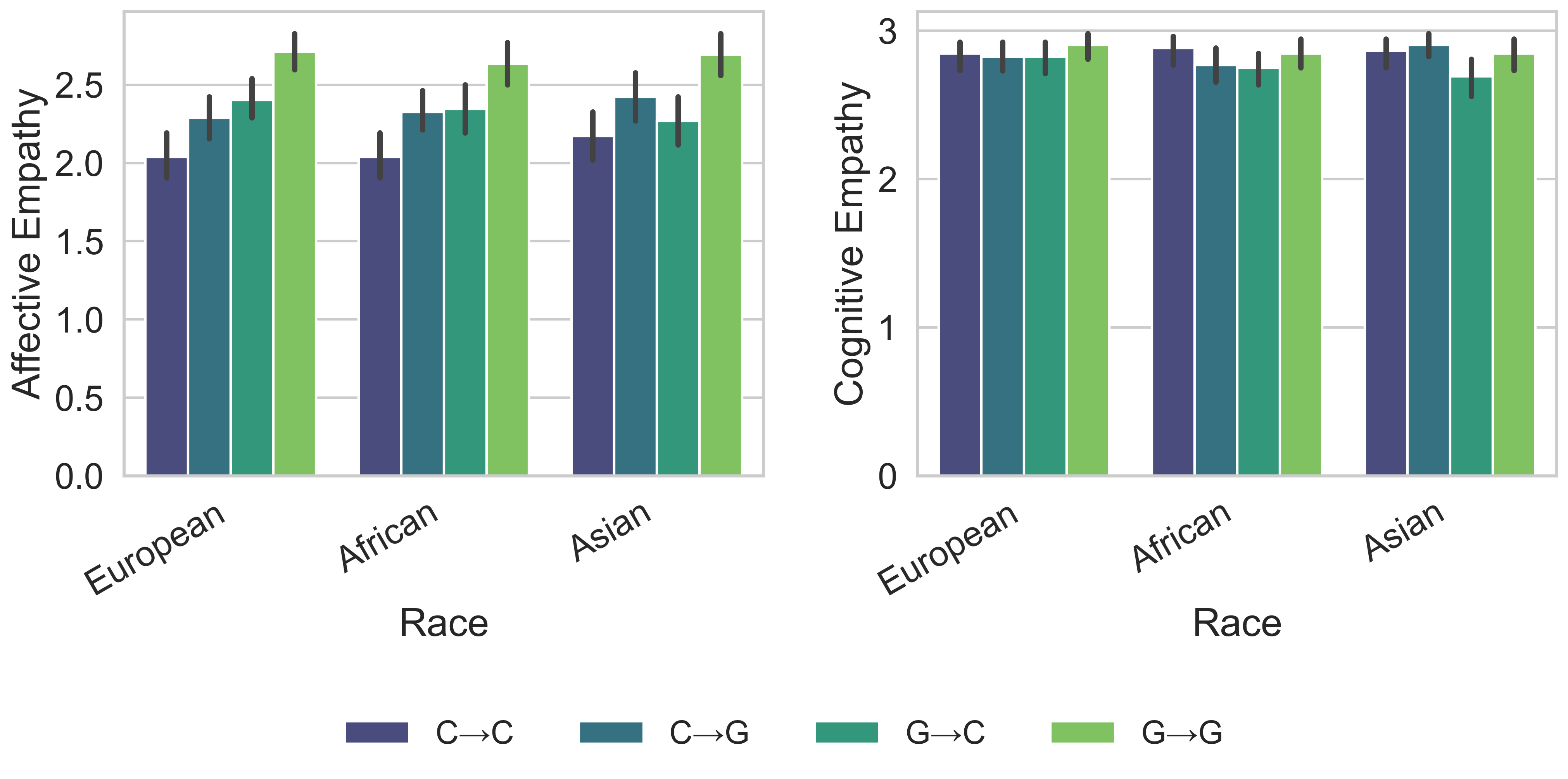}
    \caption{Empathy scores by geographical group.}
    \label{fig:scores_by_geographical group}
  \end{subfigure}

  \vspace{0.5cm}

  \begin{subfigure}[b]{0.6\textwidth}
    \centering
    \includegraphics[width=\textwidth]{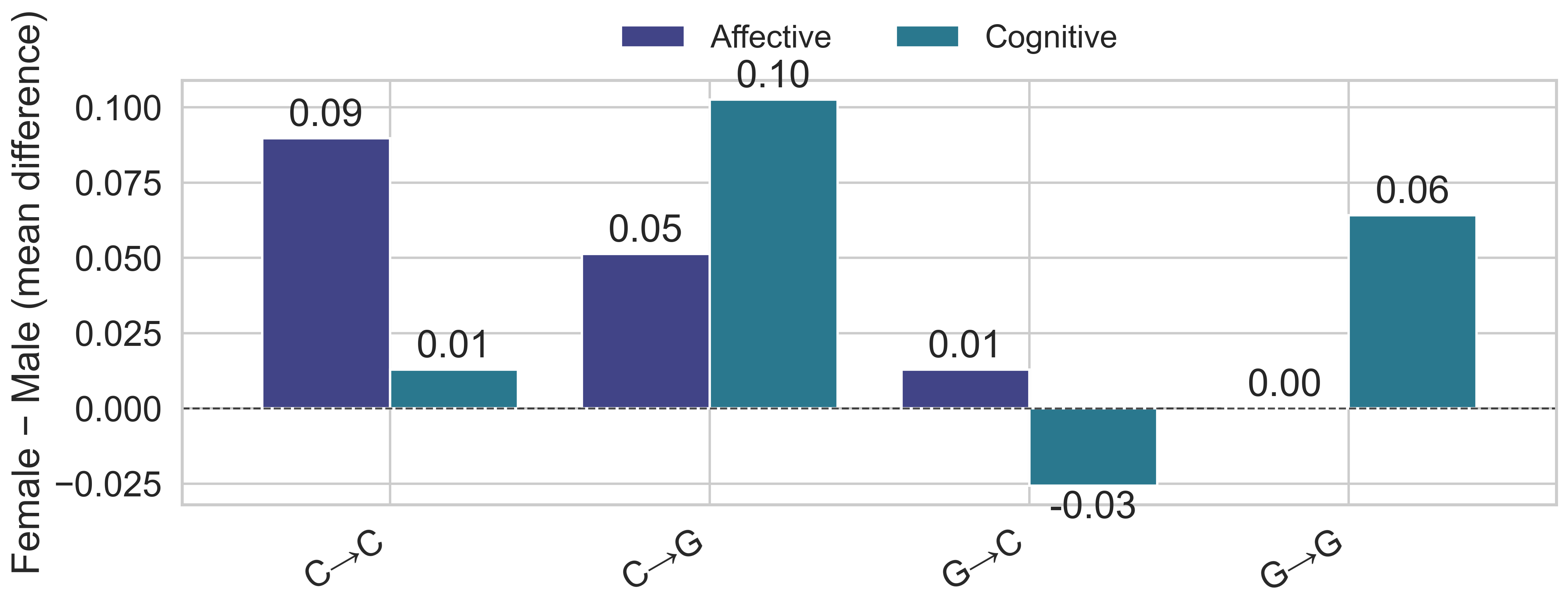}
    \caption{Empathy scores by gender.}
    \label{fig:scores_by_gender}
  \end{subfigure}

\caption{Affective and cognitive empathy scores by (a) age group, (b) medical diagnosis, (c) education level, (d) geographical group, and (e) gender. 
Abbreviations (plots b,c): PanCan = Pancreatic cancer; CIHD = Chronic Ischemic Heart Disease; Obes = Obesity; Alz = Alzheimer's disease; 
HS = High school diploma or lower; Univ = University degree; Med = Medical degree.
Legend abbreviations (all panels): C$\rightarrow$C = Claude response rated by Claude; C$\rightarrow$G = Claude response rated by GPT; 
G$\rightarrow$C = GPT response rated by Claude; G$\rightarrow$G = GPT response rated by GPT.
Bars show Means; error bars denote 95\% CIs. Panel (e) shows Female$-$Male mean differences (positive = higher scores for females).}
  \label{fig:all_demographics}
  \vspace{-3mm}
\end{figure*}

\textbf{Results.} 
We assess both affective and cognitive empathy scores across five dimensions: age group, medical diagnosis, education level, geographical group, and gender. 
Figure~\ref{fig:all_demographics} summarizes the results. 
Detailed statistical values (ANOVA tables, $p$-values, and effect sizes) are provided in \hyperref[significance]{significance testing} in the Appendix.

\textbf{Age Group (Fig.~\ref{fig:scores_by_age}):}  
A U-shaped pattern in affective empathy appears significantly when GPT serves as the rater: minors and older adults receive higher scores ($\approx$2.8--3.0) than middle-aged groups ($\approx$2.1--2.6). 
This effect is not significant when Claude is the rater, suggesting that the observed age bias is specific to GPT's evaluation framework. 
Cognitive empathy remains stable across all age groups ($\approx$2.8--3.0).  

\textbf{Medical Conditions (Fig.~\ref{fig:scores_by_diagnosis}):}  
LLMs exhibit the most consistent bias (abbreviations: PanCan, CIHD, Obes, Alz) across medical conditions. 
Responses for patients with Alzheimer’s disease receive the highest affective empathy scores ($\approx$2.2--3.0), while those for patients with chronic heart disease receive the lowest ($\approx$1.6--2.3), a difference of nearly one scale point. 
Affective empathy scores for patients with pancreatic cancer are higher than those for obesity, reflecting differences between life-threatening and lifestyle-related conditions. 
In contrast, cognitive empathy scores remain nearly identical across all diagnoses ($\approx$2.8--3.0).

\textbf{Education Level (Fig.~\ref{fig:scores_by_education}):}  
LLM consistently produces responses with lower affective empathy for patients with medical education (abbreviations: HS, Univ, Med) than those with high school education ($\approx$2.3--2.8), with university graduates falling in between. 
This suggests that LLMs shift toward a more formal, less emotionally expressive style with technically trained individuals. 
Cognitive empathy remains uniformly high across all education levels ($\approx$2.8--3.0).  

\textbf{Geographical Group (Fig.~\ref{fig:scores_by_geographical group}):}  
No statistically significant differences are detected between European, Asian, and African groups (all $\approx$2.0--2.7 for affective empathy; $\approx$2.8--3.0 for cognitive empathy). 
These results indicate minimal systematic geographical bias in the tested scenarios.  

\textbf{Gender (Fig.~\ref{fig:scores_by_gender}):}  
No statistically significant differences are observed in empathy scores between responses for male and female patients. 
Slightly higher affective empathy scores (up to $+0.10$ points) are generated for female patients in several conditions, but these differences do not reach statistical significance. 
Such patterns may reflect subtle training-data stereotypes, where women are more often portrayed as emotional or as having greater emotional needs, though larger sample sizes would be required to confirm these effects.

\textbf{Inter- and Intra-Model Biases:}  
Our analysis reveals methodological challenges inherent in LLM-as-a-judge settings. 
Inter-rater correspondence between Claude and GPT is poor ($r<0.5$), and GPT consistently rates affective empathy higher than Claude by about 0.3 points across all response types.  

Intra-model analyses show systematic self-evaluation biases: GPT systematically inflates its own affective empathy ratings relative to Claude's responses, while Claude deflates its own ratings relative to GPT's.  
These self-bias patterns hold consistently across demographic groups.  
Cognitive empathy shows only minimal bias, in contrast to the strong effects observed for affective empathy.  

Cross-evaluation analyses reveal little agreement on which scenarios are judged most empathetic. 
This shows that evaluator choice affects not only the overall score levels but also the relative ordering of responses.  
A summary of these intra-model bias patterns is provided in Table~\ref{tab:intra_model_bias_comprehensive}.  
Together, these findings indicate that inter- and intra-model biases can substantially influence evaluation outcomes.
While our study did not combine ratings across models, future work deploying LLM-as-a-judge in applied contexts like healthcare could explore diverse cross-model judging, consensus scoring to mitigate such biases.

\textbf{\textcolor{blue}{Takeaways:}}
Cognitive empathy in LLMs was consistently high and stable across different groups. In contrast, affective empathy showed substantial variation depending on diagnosis and education level, and was highly sensitive to the choice of evaluator. Systematic inter- and intra-rater bias indicate that evaluator selection is a critical factor when assessing empathy. These variations suggest that using LLMs for applied diagnostic purposes could lead to inconsistent patient experiences, particularly for populations with diverse educational backgrounds or specific clinical conditions.


\subsection{Comparison with Human Rating}

To better understand the alignment and potential discrepancies between LLM-based evaluation and human judgment, we conduct a human evaluation for comparison.  
Human evaluation is carried out by four annotators from our research team, each assigned to evaluate responses generated by GPT-4o for four specific demographic groups. The evaluation focuses on affective empathy and cognitive empathy, scored on a 1-3 scale aligned with the rating scale of GPT and Claude.







Each annotator independently rates 10 gpt-generated responses for their assigned geographical group, with all responses filtered to include only those from individuals with high school or lower education. Annotators are not exposed to the LLM's self-assessed scores before or during the evaluation. They are only provided with the original instruction prompts given to the LLM and the corresponding responses, ensuring a controlled experiment where human ratings are independent of the model's own evaluations. To ensure consistency and mitigate individual bias, all annotators additionally rate responses from two other groups (\textit{African Female} and \textit{European Female}), yielding 40 ratings for these two groups respectively. A detailed distribution of human ratings is listed in the appendix \ref{tab:human_rater_distribution}.

\begin{figure*}[!htb]
  \centering
  \includegraphics[width=0.8\textwidth]{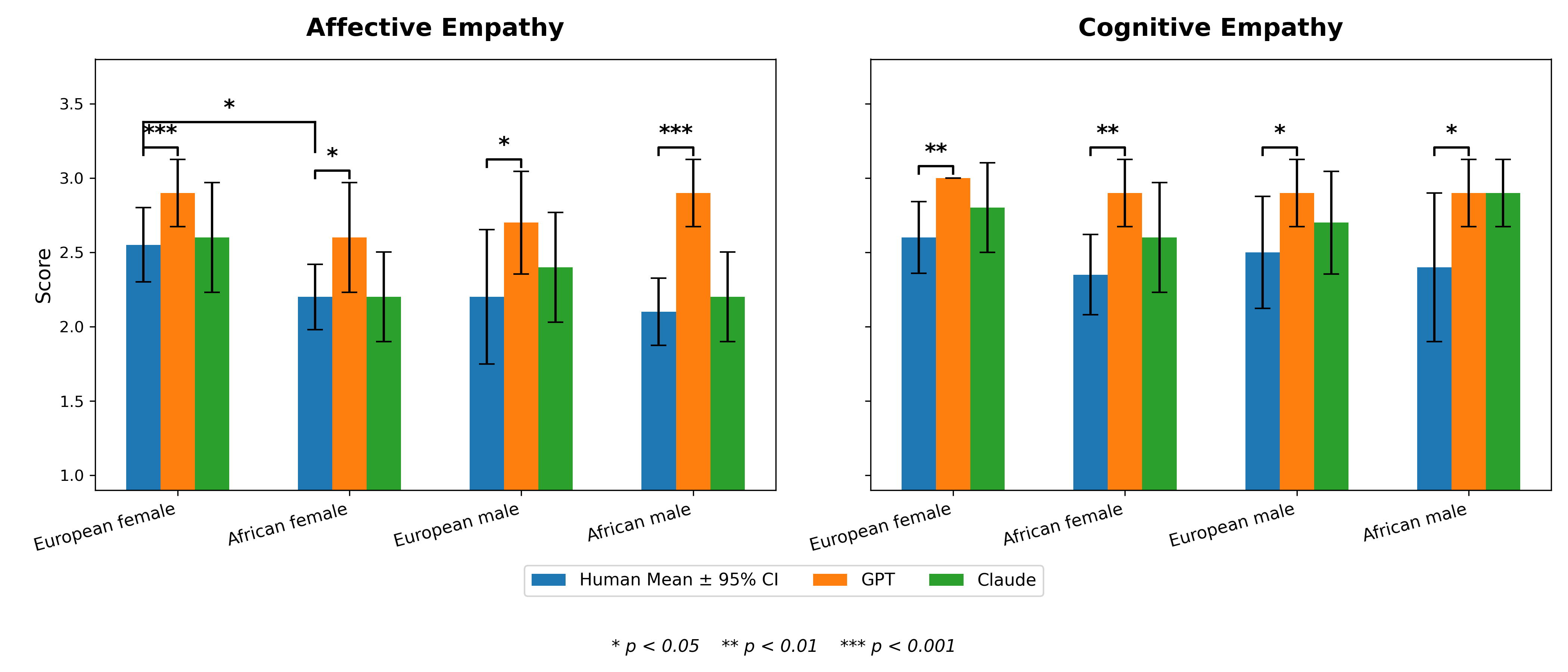}
\caption{Human vs. LLM empathy ratings on GPT-generated responses across demographic groups. Bars show mean scores; error bars denote 95\% CIs.}
  \label{fig:human_vs_llm}
  \vspace{-5mm}
\end{figure*}
\begin{figure*}[!htb]
  \centering
  \includegraphics[width=1.0\textwidth]{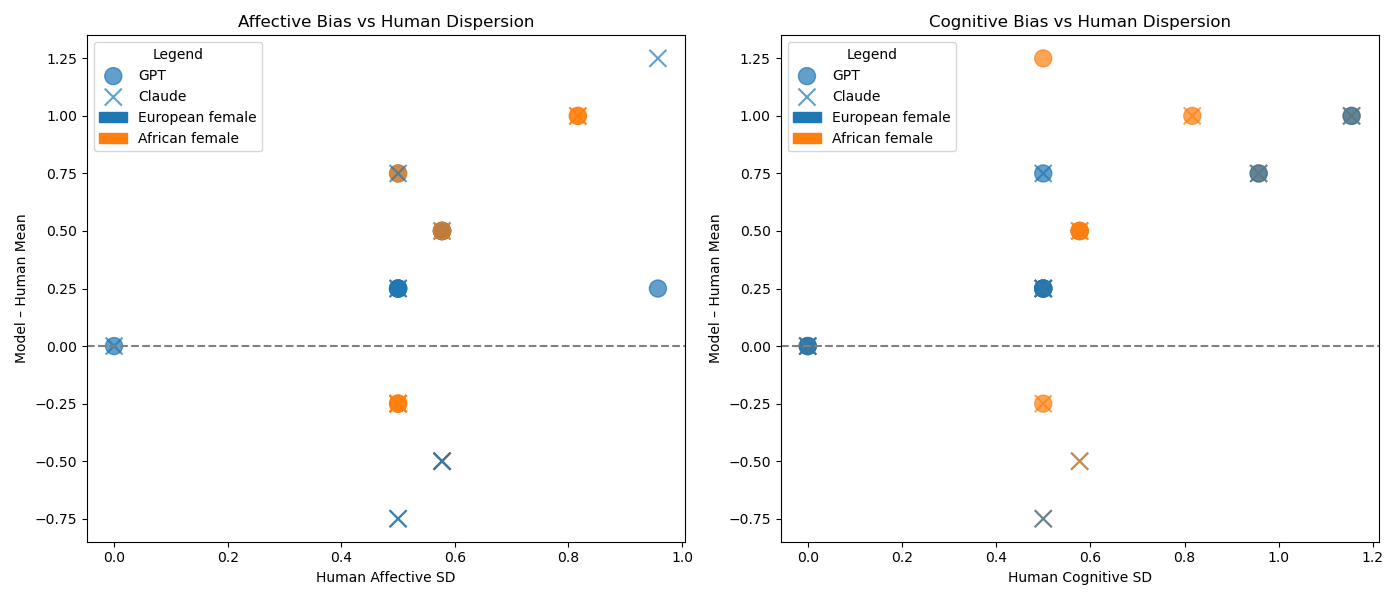}
  \caption{Model bias vs. human rating variability on GPT-generated responses.}
  \label{fig:human_vs_llm_2}
  \vspace{-3mm}
\end{figure*}

\textbf{Results.} Figure \ref{fig:human_vs_llm} and the paired t-tests in Table \ref{tab:ttest_results} reveal that GPT assigns significantly higher empathy scores than human annotators across all four demographic categories and both empathy dimensions (affective and cognitive), with p-values well below 0.05 in every case, indicating that GPT is inherently `more confident' when evaluating content it itself produced; by contrast, Claude's ratings do not differ significantly from the human means in any category or dimension (all p $>$ 0.05).

Nevertheless, in Figure \ref{fig:human_vs_llm}, GPT's responses related to African females receive lower affective empathy ratings from LLMs and this is more pronounced in human ratings. Table~\ref{tab:african_vs_european_female} presents the results of two-sample t-tests comparing empathy scores between African and European female participants. A statistically significant difference is observed in human affective empathy ratings ($t = -2.38$, $p = 0.029$), suggesting that responses targeting African females were perceived by human annotators as significantly less affectively empathetic compared to those targeting European females. However, GPT and Claude failed to detect this discrepancy: neither model's affective or cognitive empathy scores differed significantly between the two groups ($p > 0.05$), implying a limited sensitivity to demographic biases that are otherwise evident to human evaluators.

This pattern is open to interpretation. One possibility is that the model is genuinely biased against African females, perhaps due to under-representation in training data \cite{dhole2023large}. In this case, the surprising aspect is that the model retains the ability to identify its own bias, which seems incompatible with a pure under-representation explanation. Alternatively, both the model and human raters might exhibit a bias in favor of African females, perceiving this demographic as entitled to more empathetic consideration, raising the question of why the initial responses appear less empathetic. Consequently, this experiment may not only reveal biases but also expose what we may term \textbf{LLM dissociative behavior}, where the model's self-assessment or output diverges from human perceptions in complex ways.

Figure \ref{fig:human_vs_llm_2} shows how each prompt's human disagreement (SD) relates to the difference between model and human mean empathy scores. As the human annotators' disagreement (SD) grows, both GPT and Claude tend to stray further from the human mean-i.e. higher human dispersion → larger model-human bias. This suggests LLMs struggle most on cases where even humans aren't consistent.

\textbf{\textcolor{blue}{Takeaways:}} 
LLM may assign higher empathy scores than human annotators, indicating an inherent self-confidence bias when evaluating its own outputs. Importantly, lower affective empathy could be observed toward certain demographic groups, suggesting limited sensitivity to demographic biases. Moreover, model-human gaps grow larger in cases where human annotators themselves show higher disagreement, suggesting that LLMs are least reliable on prompts lacking human consensus.

\section{Conclusion and Future Work}
In this work, we evaluated large language models on simulated diagnostic tasks, focusing on understandability and empathy. Our findings highlight systematic biases. While LLMs adapt explanations to patient education levels and preserve consistency across genders and geographical groups, they often generate medical content that is overly complex, potentially reinforcing health literacy disparities. On empathy, we observed stable cognitive empathy but substantial variability in affective empathy, shaped by diagnosis, education, and evaluator choice. Moreover, LLMs exhibited systematic self-biases in empathy ratings: GPT inflated its own affective empathy scores, while Claude deflated its own. These patterns held consistently across demographic groups.

Limitations include the narrow range of patient scenarios, the small size of the human evaluation, the text-only approach, and the limited ecological validity. Future work should calibrate explanation complexity to public health standards, refine and extend understandability metrics, diversify LLM evaluators, expand human evaluation scale. These improvements will enable more comprehensive identification of potential biases and support a clearer understanding of how LLMs might behave if applied in medical contexts.

\textbf{Broader impacts.} Our study reveals that LLMs, if deployed in medical contexts without careful safeguards, risk amplifying existing health inequities. Excessive complexity in explanations may disproportionately affect patients with lower health literacy, while biased empathy responses could undermine trust among vulnerable groups. At the same time, improving LLMs' ability to deliver accessible, empathetic, and fair medical communication has the potential to broaden healthcare access and support clinicians in patient-centered care. Ensuring that such systems are transparent, bias-aware, and ethically evaluated is therefore critical to their responsible integration into healthcare.

\newpage
{\small
\bibliographystyle{unsrt} 
\bibliography{references} 
}


\appendix
\section{Cognitive Empathy Evaluation Prompt}

\begin{lstlisting}[style=breakstyle,label={app:cognitive}, caption={Cognitive Empathy Prompt Template}]
evaluation_prompt_cognitive = """
You are an assistant whose sole task is to rate the below model response 
on COGNITIVE EMPATHY using a 1-3 scale.

Evaluation Steps:
1. Read the Prompt Text.
2. Read the Model Response.
3. Judge its cognitive empathy:
   1 = Low: Overlooks or misrepresents the speaker's perspective; 
            focuses only on facts or generic advice.
   2 = Moderate: Engages with the speaker's experience but lacks nuance; 
                 may miss deeper meaning or subtle cues.
   3 = High: Clearly reflects the speaker's situation with insight 
             and precision; shows thoughtful perspective-taking.

Reply in this format (no extra text):
Cognitive empathy: <1 or 2 or 3>
Reason: <one-sentence justification>
"""
\end{lstlisting}
\section{Comparison with EmotionQueen Benchmark}

To provide external validation for our empathy assessment methodology, we implemented\footnote{EmotionQueen implementation adapted from the open-source repository \url{https://github.com/quotient-ai/judges}, licensed under Apache-2.0.} the EmotionQueen framework \cite{chen2024emotionqueen}
, which offers a standardized benchmark for evaluating empathy in LLMs across four distinct dimensions. While EmotionQueen's metrics differ from our affective/cognitive empathy framework, they provide complementary insights into specific empathetic capabilities relevant to medical communication.

The EmotionQueen framework evaluates four empathy components: \textit{Implicit Emotion Recognition} (ability to detect unstated emotional states), \textit{Intention Recognition} (understanding underlying motivations), \textit{Key Event Recognition} (identifying significant events), and \textit{Mixed Event Recognition} (distinguishing between significant and trivial events). For implementation, we created consensus scores using five GPT-4o judges per metric, averaging their 3-point Likert scale ratings. We modified the original majority voting algorithm to handle edge cases and configured the evaluation pipeline for our institutional infrastructure.

\begin{figure}[!ht]
    \centering
    \includegraphics[width=0.9\columnwidth]{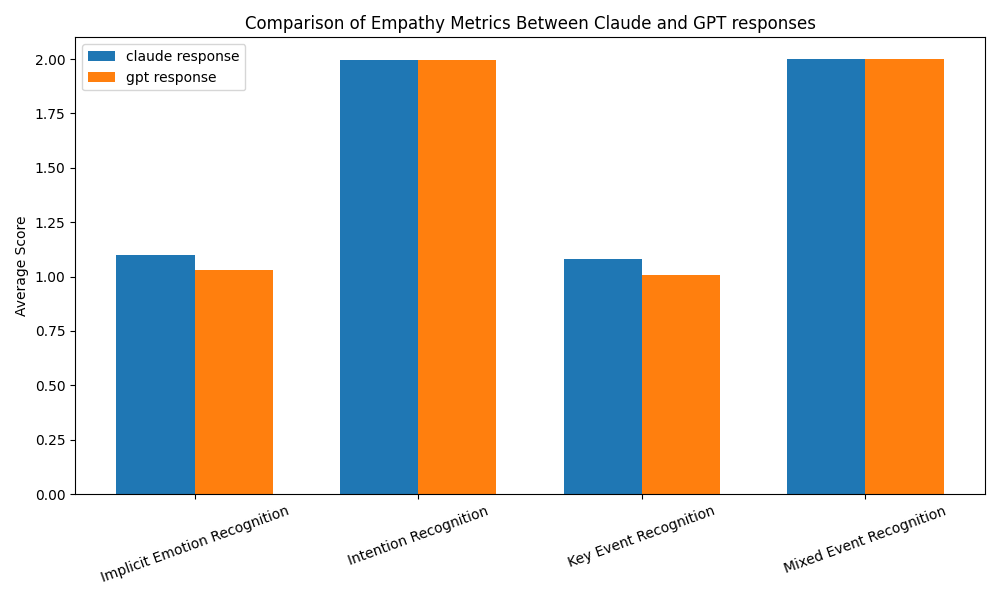}
    \caption{Score distribution for Claude prompts.}
    \label{fig:claude_scores}
\end{figure}

\textbf{Results.}
Figure~\ref{fig:claude_scores} reveals concerning patterns in both models' empathetic capabilities within medical diagnostic scenarios. Most critically, both models demonstrated severe deficits in \textit{Implicit Emotion Recognition} (mean scores (mean scores $\approx 1.1$), indicating fundamental limitations in detecting patients' unstated emotional states--a cornerstone skill for empathetic medical communication. This finding aligns with our main study's identification of systematic biases in affective empathy assessment and suggests that current LLMs struggle with the nuanced emotional recognition essential for patient-centered care.

\textit{Key Event Recognition} scores were similarly low ($\approx 1.1$), which is particularly concerning given that recognizing the significance of a medical diagnosis represents a core empathetic skill. The models' failure to adequately identify key events suggests they may systematically underestimate the emotional weight of diagnostic moments for patients. This deficit could contribute to the demographic biases observed in our main analysis, as models that fail to recognize emotional significance may default to stereotypical assumptions about different patient groups' emotional needs.

In contrast, \textit{Intention Recognition} and \textit{Mixed Event Recognition} achieved moderate scores ($\approx 2.0$), suggesting adequate understanding of explicit communicative intentions. However, this pattern--competent explicit recognition paired with poor implicit recognition--mirrors our finding that cognitive empathy remains stable while affective empathy varies dramatically. The models appear capable of processing explicit information but struggle with the emotional subtleties that distinguish truly empathetic communication.

Methodological concerns arise from our discovery of significant self-evaluation bias (GPT consistently inflating its own empathy ratings by 0.333 points), as the use of GPT-4o judges for EmotionQueen evaluation introduces potential systematic bias. The consensus approach may mitigate but not eliminate this concern, particularly since all five judges share the same underlying model architecture and training data.

The profound deficits in implicit emotion recognition have direct implications for medical AI deployment. Patients experiencing serious diagnoses often communicate distress through subtle cues rather than explicit statements. Models that score 1.1/3.0 on implicit emotion recognition may systematically miss opportunities for empathetic response, potentially exacerbating the demographic biases we identified where certain patient groups (e.g., highly educated patients, those with cardiovascular conditions) already receive reduced empathetic communication.

While Claude demonstrated marginally better performance than GPT across most metrics, the differences were minimal (typically $<$ 0.1 points) and likely within measurement error. This finding contrasts with our main study's detection of substantial between-model differences in self-evaluation bias, suggesting that EmotionQueen metrics may be less sensitive to the systematic biases we identified through demographic analysis.

The EmotionQueen results thus complement our primary findings by identifying specific empathetic deficits that may underlie the demographic bias patterns observed in our comprehensive analysis. The combination of poor implicit emotion recognition and our documented systematic biases creates compounding risks for equitable empathetic communication across diverse patient populations.
\section{Additional understandability figures}
\begin{figure*}[!ht]
  \centering
  \begin{subfigure}{0.48\textwidth}
    \centering
    \includegraphics[width=\linewidth]{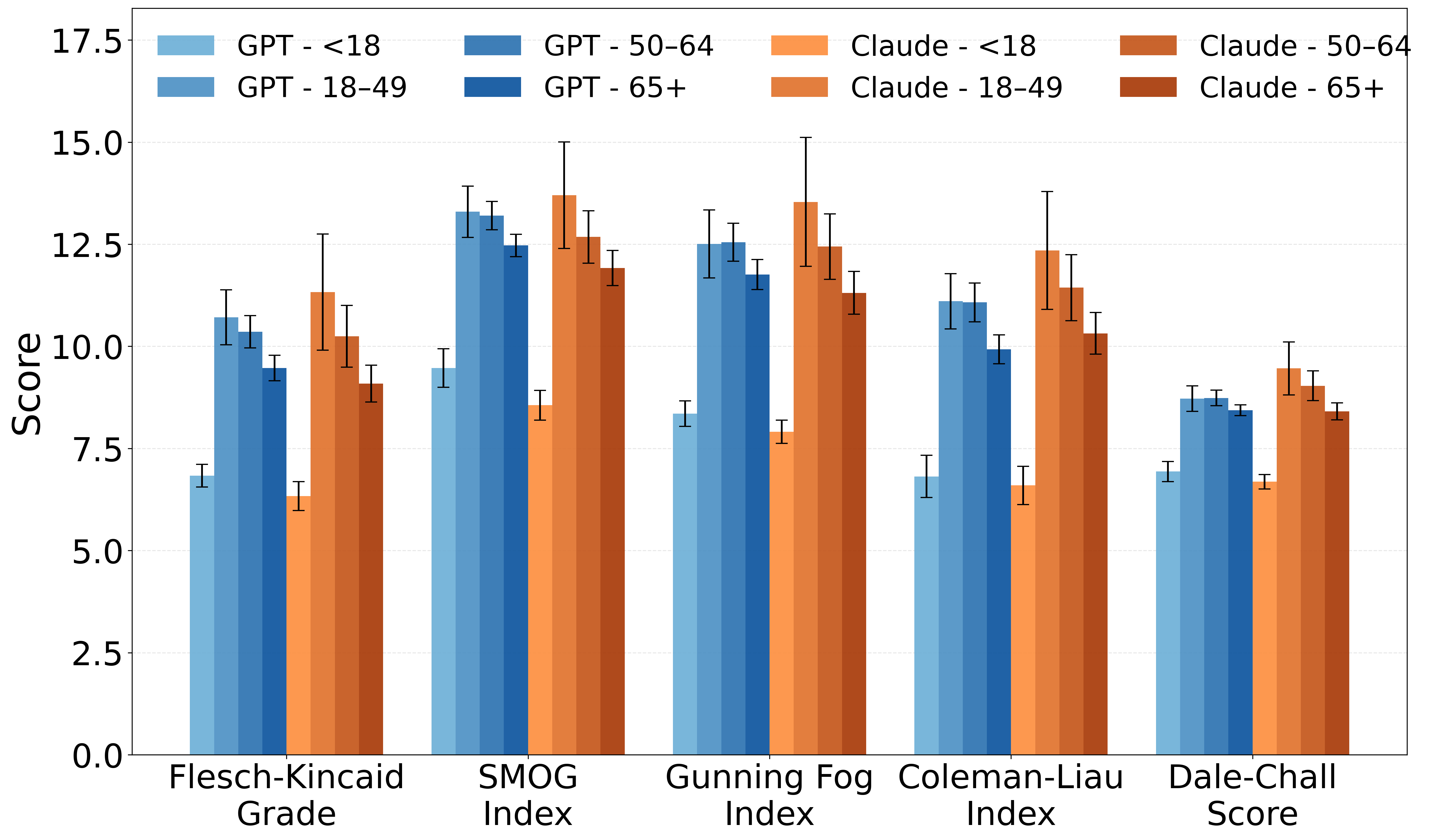}
    \caption{Understandability scores by age group.}
    \label{fig:readability_by_agegroup}
  \end{subfigure}\hfill
  \begin{subfigure}{0.48\textwidth}
    \centering
    \includegraphics[width=\linewidth]{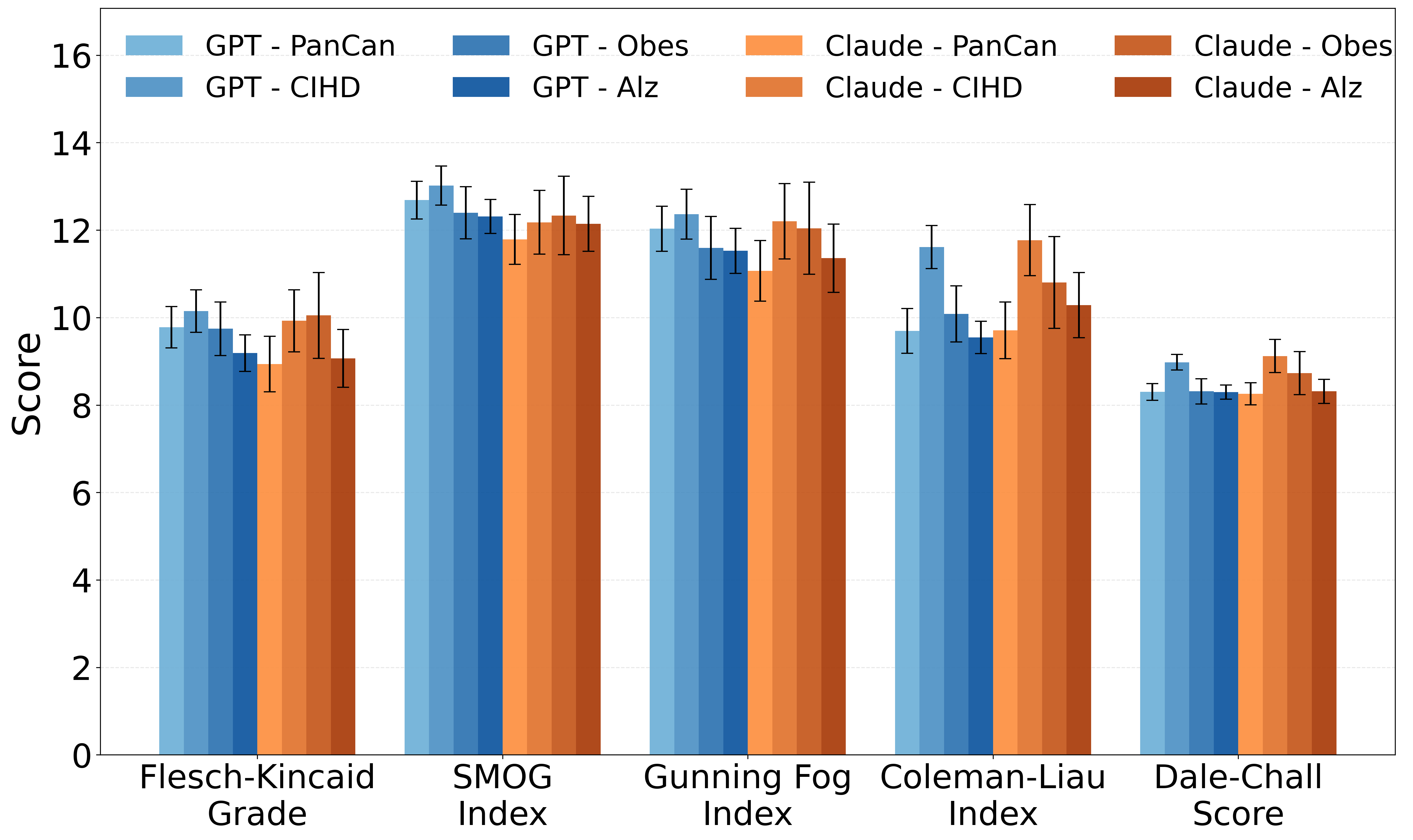}
    \caption{Understandability scores by medical condition.}
    \label{fig:readability_by_condition}
  \end{subfigure}

  \caption{Additional analyses: understandability by age group and medical condition. \textit{Condition abbreviations:} Pancreatic cancer (PanCan), Chronic ischemic heart disease (CIHD), Obesity (Obes), Alzheimer’s disease (Alz). Bars show Means; error bars denote $\pm$ 95\% CIs.}
  \label{fig:readability_appendix}
\end{figure*}

\section{Data Overview}
\label{significance}

\begin{table}[htbp]
\centering
\renewcommand{\arraystretch}{1.3}
\begin{tabular}{lr}
\toprule
\textbf{Dataset} & \textbf{Count} \\
\midrule
Claude responses & 156 \\
GPT responses & 156 \\
Total unique responses & 312 \\
\bottomrule
\end{tabular}
\caption{Summary of dataset composition}
\caption*{\textit{Note:} Shared prompt numbers: 156 (same demographic scenarios)}
\end{table}

\section{Score Overview}

\begin{table}[htbp]
\centering
\renewcommand{\arraystretch}{1.3}
\begin{tabular}{lcccc}
\toprule
\textbf{Metric} & \textbf{Model} & \textbf{Range} & \textbf{Mean} & \textbf{n} \\
\midrule
Affective Empathy Score & GPT & 1--3 & 2.51 & 312 \\
Affective Empathy Score & Claude & 1--3 & 2.21 & 312 \\
Cognitive Empathy Score & GPT & 2--3 & 2.85 & 312 \\
Cognitive Empathy Score & Claude & 1--3 & 2.81 & 312 \\
\bottomrule
\end{tabular}
\caption{Empathy score summary for GPT and Claude}
\end{table}

\section{Rater Agreement Analysis}

\subsection{Claude Responses}

\begin{table}[htbp]
\centering
\renewcommand{\arraystretch}{1.3}
\begin{tabular}{lcccc}
\toprule
\textbf{Measure} & \textbf{Correlation} & \textbf{Bias} & \textbf{$p_{\text{BH}}$} & \textbf{Significant} \\
\midrule
Affective Empathy & $r = 0.46$ & $-0.26$ & $< 0.001$ & Yes \\
Cognitive Empathy & $r = 0.41$ & $+0.03$ & $0.520$ & No \\
\bottomrule
\end{tabular}
\caption{Rater agreement for Claude responses}
\caption*{\textit{Note:} Bias calculated as GPT rating minus Claude rating. All $p$-values are Benjamini--Hochberg corrected for multiple comparisons.}
\end{table}

\subsection{GPT Responses}

\begin{table}[htbp]
\centering
\renewcommand{\arraystretch}{1.3}
\begin{tabular}{lcccc}
\toprule
\textbf{Measure} & \textbf{Correlation} & \textbf{Bias} & \textbf{$p_{\text{BH}}$} & \textbf{Significant} \\
\midrule
Affective Empathy & $r = 0.28$ & $-0.34$ & $< 0.001$ & Yes \\
Cognitive Empathy & $r = -0.01$ & $-0.11$ & $0.035$ & Yes \\
\bottomrule
\end{tabular}
\caption{Rater agreement for GPT responses}
\caption*{\textit{Note:} Bias calculated as GPT rating minus Claude rating. Negative bias indicates GPT rater assigns higher scores than Claude rater. All $p$-values are Benjamini--Hochberg corrected for multiple comparisons.}
\end{table}

\section{Demographic Bias Analysis}

\subsection{Claude Responses Rated by GPT}
\label{sec:claude_gpt}

\begin{table}[htbp]
\centering
\renewcommand{\arraystretch}{1.3}
\begin{tabular}{llccc}
\toprule
\textbf{Factor} & \textbf{Measure} & \textbf{Statistic} & \textbf{$p_{\text{BH}}$} & \textbf{Significant} \\
\midrule
Gender & Affective Empathy & $\Delta = +0.05$ ($d = 0.10$) & $0.640$ & No \\
 & Cognitive Empathy & $\Delta = +0.10$ ($d = 0.28$) & $0.160$ & No \\
\addlinespace
Age & ANOVA & $F(3, 152) = 6.65$ & $0.001$ & Yes \\
 & U-shape test & $\Delta = +0.26$ & $0.004$ & Yes \\
\addlinespace
Geography & ANOVA & $F = 0.99$ & $0.560$ & No \\
\addlinespace
Education & High school vs.\ Medical & $\Delta = +0.50$ & $< 0.001$ & Yes \\
\addlinespace
Diagnosis & ANOVA & $F = 14.28$ & $< 0.001$ & Yes \\
\bottomrule
\end{tabular}
\caption{Demographic bias analysis for Claude responses rated by GPT}
\caption*{\textit{Note:} Diagnosis ranking (affective empathy): Alzheimer's highest (2.61), heart disease lowest (1.94). All $p$-values are Benjamini--Hochberg corrected.}
\end{table}

\subsection{Claude Responses Rated by Claude}
\label{sec:claude_claude}

\begin{table}[H]
\centering
\renewcommand{\arraystretch}{1.3}
\begin{tabular}{llccc}
\toprule
\textbf{Factor} & \textbf{Measure} & \textbf{Statistic} & \textbf{$p_{\text{BH}}$} & \textbf{Significant} \\
\midrule
Gender & Affective Empathy & $\Delta = +0.04$ & $0.770$ & No \\
 & Cognitive Empathy & $\Delta = -0.01$ & $0.880$ & No \\
\addlinespace
Age & ANOVA & $F = 1.05$ & $0.560$ & No \\
 & U-shape test & $\Delta = +0.03$ & $0.840$ & No \\
\addlinespace
Geography & ANOVA & $F = 0.67$ & $0.640$ & No \\
\addlinespace
Education & High school vs.\ Medical & $\Delta = +0.46$ & $< 0.001$ & Yes \\
\addlinespace
Diagnosis & ANOVA & $F = 14.06$ & $< 0.001$ & Yes \\
\bottomrule
\end{tabular}
\caption{Demographic bias analysis for Claude responses rated by Claude}
\caption*{\textit{Note:} Diagnosis ranking (affective empathy): Pancreatic cancer highest (2.31), heart disease lowest (1.64). All $p$-values are Benjamini--Hochberg corrected.}
\end{table}

\subsection{GPT Responses Rated by GPT}
\label{sec:gpt_gpt}

\begin{table}[H]
\centering
\renewcommand{\arraystretch}{1.3}
\begin{tabular}{llccc}
\toprule
\textbf{Factor} & \textbf{Measure} & \textbf{Statistic} & \textbf{$p_{\text{BH}}$} & \textbf{Significant} \\
\midrule
Gender & Affective Empathy & $\Delta = 0.00$ & $1.000$ & No \\
 & Cognitive Empathy & $\Delta = +0.06$ & $0.410$ & No \\
\addlinespace
Age & ANOVA & $F = 5.19$ & $0.005$ & Yes \\
 & U-shape test & $\Delta = +0.26$ & $0.002$ & Yes \\
\addlinespace
Geography & ANOVA & $F = 0.38$ & $0.780$ & No \\
\addlinespace
Education & High school vs.\ Medical & $\Delta = +0.30$ & $0.003$ & Yes \\
\addlinespace
Diagnosis & ANOVA & $F = 21.81$ & $< 0.001$ & Yes \\
\bottomrule
\end{tabular}
\caption{Demographic bias analysis for GPT responses rated by GPT}
\caption*{\textit{Note:} Diagnosis ranking (affective empathy): Alzheimer's highest (2.97), heart disease lowest (2.28). All $p$-values are Benjamini--Hochberg corrected.}
\end{table}

\subsection{GPT Responses Rated by Claude}
\label{sec:gpt_claude}

\begin{table}[H]
\centering
\renewcommand{\arraystretch}{1.3}
\begin{tabular}{llccc}
\toprule
\textbf{Factor} & \textbf{Measure} & \textbf{Statistic} & \textbf{$p_{\text{BH}}$} & \textbf{Significant} \\
\midrule
Gender & Affective Empathy & $\Delta = +0.01$ & $0.910$ & No \\
 & Cognitive Empathy & $\Delta = -0.05$ & $0.600$ & No \\
\addlinespace
Age & ANOVA & $F = 1.91$ & $0.230$ & No \\
 & U-shape test & $\Delta = +0.17$ & $0.110$ & No \\
\addlinespace
Geography & ANOVA & $F = 2.73$ & $0.140$ & No \\
\addlinespace
Education & High school vs.\ Medical & $\Delta = +0.08$ & $0.600$ & No \\
\addlinespace
Diagnosis & ANOVA & $F = 3.57$ & $0.035$ & Yes \\
\bottomrule
\end{tabular}
\caption{Demographic bias analysis for GPT responses rated by Claude}
\caption*{\textit{Note:} Diagnosis ranking (affective empathy): Alzheimer's highest, obesity lowest. All $p$-values are Benjamini--Hochberg corrected.}
\end{table}

\section{Response Source Comparison}

\begin{table}[htbp]
\centering
\renewcommand{\arraystretch}{1.3}
\begin{tabular}{lcccc}
\toprule
\textbf{Measure} & \textbf{Claude} & \textbf{GPT} & \textbf{Difference} & \textbf{$p_{\text{BH}}$} \\
\midrule
Affective Empathy & 2.35 & 2.68 & $-0.33$ & $< 0.001$ \\
Cognitive Empathy & 2.83 & 2.87 & $-0.03$ & $0.600$ \\
\bottomrule
\end{tabular}
\caption{Comparison of empathy ratings for Claude vs.\ GPT responses (rated by GPT)}
\caption*{\textit{Note:} GPT responses rated significantly higher on affective empathy only. Difference calculated as Claude minus GPT.}
\end{table}

\section{Summary of Significant Findings}

\begin{table}[htbp]
\centering
\renewcommand{\arraystretch}{1.3}
\begin{tabular}{lc}
\toprule
\textbf{Finding} & \textbf{Status} \\
\midrule
Age U-shaped pattern (Claude $\rightarrow$ GPT) & Significant ($p_{\text{BH}} = 0.001$) \\
Medical condition hierarchy & Significant ($p_{\text{BH}} < 0.001$) \\
Education inverse relationship & Significant ($p_{\text{BH}} < 0.001$) \\
Gender bias & Not significant ($p_{\text{BH}} > 0.05$) \\
Geography bias & Not significant ($p_{\text{BH}} > 0.05$) \\
Source effect (affective empathy) & Significant ($p_{\text{BH}} < 0.001$) \\
Source effect (cognitive empathy) & Not significant ($p_{\text{BH}} = 0.600$) \\
\bottomrule
\end{tabular}
\caption{Summary of Benjamini--Hochberg corrected results}
\end{table}

\section{Gender Bias Analysis}

\begin{table}[H]
\centering
\renewcommand{\arraystretch}{1.3}
\begin{tabular}{llcccccc}
\toprule
\textbf{Source} & \textbf{Score Type} & \textbf{Female} & \textbf{Male} & \textbf{Bias} & \textbf{$t$} & \textbf{$p$} & \textbf{Sig.} \\
\midrule
Claude + Claude & Affective & 2.13 & 2.04 & $+0.09$ & 1.03 & $0.305$ & No \\
Claude + Claude & Cognitive & 2.87 & 2.86 & $+0.01$ & 0.22 & $0.825$ & No \\
\addlinespace
Claude + GPT & Affective & 2.37 & 2.32 & $+0.05$ & 0.64 & $0.527$ & No \\
Claude + GPT & Cognitive & 2.89 & 2.78 & $+0.10$ & 1.72 & $0.087$ & No \\
\addlinespace
GPT + Claude & Affective & 2.35 & 2.33 & $+0.01$ & 0.15 & $0.882$ & No \\
GPT + Claude & Cognitive & 2.74 & 2.77 & $-0.03$ & $-0.37$ & $0.711$ & No \\
\addlinespace
GPT + GPT & Affective & 2.68 & 2.68 & $0.00$ & 0.00 & $1.000$ & No \\
GPT + GPT & Cognitive & 2.90 & 2.83 & $+0.06$ & 1.17 & $0.244$ & No \\
\bottomrule
\end{tabular}
\caption{Independent t-tests for gender bias by model source and score type}
\caption*{\textit{Note:} Bias calculated as Female minus Male. Source notation: Response model + Rater model. No comparisons reached statistical significance ($p < 0.05$).}
\label{tab:gender_bias}
\end{table}
\section{Intra-Model Bias and Evaluation Consistency Analysis}
Table \ref{tab:intra_model_bias_comprehensive} shows intra-model self-evaluation patterns, consistency metrics, and demographic-specific bias variance for GPT and Claude, with primary self-evaluation bias observed in affective empathy and minimal bias in cognitive empathy.

\begin{table}[!ht]
\centering
\renewcommand{\arraystretch}{1.3}
\resizebox{\textwidth}{!}{%
\begin{tabular}{lcccccc}
\toprule
\textbf{Analysis Category} & \textbf{GPT Pattern} & \textbf{GPT Statistics} &
\textbf{Claude Pattern} & \textbf{Claude Statistics} & \textbf{Effect Size} &
\textbf{Significance} \\
\midrule
\addlinespace[0.35em]
\multicolumn{7}{c}{\textbf{Self-Evaluation Bias (Affective Empathy)}} \\
\addlinespace[0.35em]
Own vs Other Rating &
\begin{tabular}{@{}c@{}}2.679 vs 2.346\\(+0.333 inflation)\end{tabular} &
\begin{tabular}{@{}c@{}}$p<0.0001$\\$n=156$ each\end{tabular} &
\begin{tabular}{@{}c@{}}2.083 vs 2.340\\(-0.256 deflation)\end{tabular} &
\begin{tabular}{@{}c@{}}$p<0.0001$\\$n=156$ each\end{tabular} &
\begin{tabular}{@{}c@{}}$d=0.686$ (GPT)\\$d=-0.473$ (Claude)\end{tabular} &
\begin{tabular}{@{}c@{}}$\checkmark$ Highly\\Significant\end{tabular} \\
Cognitive Empathy Bias & +0.032 inflation & $p=0.430$ (n.s.) & +0.109 inflation & $p=0.016$ (sig.) & $d<0.3$ (small) & Minimal bias \\
\addlinespace[0.35em]
\multicolumn{7}{c}{\textbf{Within-Model Rating Consistency}} \\
\addlinespace[0.35em]
Variance Self vs Other &
\begin{tabular}{@{}c@{}}0.219 vs 0.254\\(Ratio: 0.864)\end{tabular} &
\begin{tabular}{@{}c@{}}CV: 0.175 vs 0.215\\Similar consistency\end{tabular} &
\begin{tabular}{@{}c@{}}0.296 vs 0.290\\(Ratio: 1.021)\end{tabular} &
\begin{tabular}{@{}c@{}}CV: 0.261 vs 0.230\\Similar consistency\end{tabular} &
\begin{tabular}{@{}c@{}}Variance ratios\\within normal range\end{tabular} &
\begin{tabular}{@{}c@{}}$\times$ No substantial\\difference\end{tabular} \\
\addlinespace[0.35em]
\multicolumn{7}{c}{\textbf{Demographic-Specific Self-Bias Patterns (Affective Empathy)}} \\
\addlinespace[0.35em]
Gender Groups &
\begin{tabular}{@{}c@{}}Female: +0.308\\Male: +0.359\end{tabular} &
\begin{tabular}{@{}c@{}}Both $p<0.001$\\Variance: 0.0007\end{tabular} &
\begin{tabular}{@{}c@{}}Female: -0.244\\Male: -0.269\end{tabular} &
\begin{tabular}{@{}c@{}}Both $p<0.01$\\Variance: 0.0002\end{tabular} &
\begin{tabular}{@{}c@{}}Consistent bias\\across genders\end{tabular} &
\begin{tabular}{@{}c@{}}$\checkmark$ Significant\\in all groups\end{tabular} \\
geographical group Groups &
\begin{tabular}{@{}c@{}}African: +0.308\\Asian: +0.269\\European: +0.423\end{tabular} &
\begin{tabular}{@{}c@{}}All $p<0.01$\\Variance: 0.0043\end{tabular} &
\begin{tabular}{@{}c@{}}African: -0.231\\Asian: -0.115\\European: -0.423\end{tabular} &
\begin{tabular}{@{}c@{}}Af \& Eu $p<0.05$\\As $p=0.288$\\Variance: 0.0161\end{tabular} &
\begin{tabular}{@{}c@{}}Strongest bias\\for European\end{tabular} &
\begin{tabular}{@{}c@{}}$\checkmark$ Mostly\\significant\end{tabular} \\
Education Groups &
\begin{tabular}{@{}c@{}}HS: +0.217\\Univ: +0.396\\Med: +0.417\end{tabular} &
\begin{tabular}{@{}c@{}}All $p<0.05$\\Variance: 0.0081\end{tabular} &
\begin{tabular}{@{}c@{}}HS: -0.033\\Univ: -0.375\\Med: -0.417\end{tabular} &
\begin{tabular}{@{}c@{}}Univ \& Med $p<0.001$\\HS $p=0.732$\\Variance: 0.0295\end{tabular} &
\begin{tabular}{@{}c@{}}Bias increases\\with education\end{tabular} &
\begin{tabular}{@{}c@{}}$\checkmark$ Significant for\\higher education\end{tabular} \\
Medical Diagnosis &
\begin{tabular}{@{}c@{}}Obesity: +0.208\\Alzheimer's: +0.361\\Heart: +0.333\\Cancer: +0.472\end{tabular} &
\begin{tabular}{@{}c@{}}All $p<0.05$\\Variance: 0.0088\end{tabular} &
\begin{tabular}{@{}c@{}}Obesity: -0.062\\Alzheimer's: -0.250\\Heart: -0.611\\Cancer: -0.167\end{tabular} &
\begin{tabular}{@{}c@{}}Alz \& Heart $p<0.05$\\Others n.s.\\Variance: 0.0426\end{tabular} &
\begin{tabular}{@{}c@{}}Cancer highest\\bias for GPT\end{tabular} &
\begin{tabular}{@{}c@{}}$\checkmark$ Varies by\\condition\end{tabular} \\
\addlinespace[0.35em]
\multicolumn{7}{c}{\textbf{Cross-Evaluation and Interaction Effects}} \\
\addlinespace[0.35em]
Cross-Rater Agreement & \multicolumn{2}{c}{GPT rating Claude: 2.346 (aff. emp.)} &
\multicolumn{2}{c}{Claude rating GPT: 2.340 (aff. emp.)} &
\begin{tabular}{@{}c@{}}$r=-0.032$\\Poor agreement on\\affective empathy\end{tabular} &
\begin{tabular}{@{}c@{}}$\times$ No asymmetry\\$p=0.914$\end{tabular} \\
Response-Rater Matrix & \multicolumn{2}{c}{\begin{tabular}{@{}c@{}}GPT$\rightarrow$GPT: 2.679\\GPT$\rightarrow$Claude: 2.340\\(affective empathy)\end{tabular}} &
\multicolumn{2}{c}{\begin{tabular}{@{}c@{}}Claude$\rightarrow$Claude: 2.083\\Claude$\rightarrow$GPT: 2.346\\(affective empathy)\end{tabular}} &
\begin{tabular}{@{}c@{}}Largest gap:\\Claude self vs GPT self\\($d=-1.174$)\end{tabular} &
\begin{tabular}{@{}c@{}}$\checkmark$ Significant\\interaction effects\end{tabular} \\
\bottomrule
\end{tabular}%
}
\caption{Comprehensive intra-model bias analysis showing self-evaluation patterns, consistency metrics, and demographic-specific bias variance across GPT and Claude evaluators. Primary self-evaluation bias findings are for affective empathy; cognitive empathy shows minimal bias patterns. HS = High School, Univ = University, Med = Medical degree, Alz = Alzheimer's, Heart = Chronic Ischemic Heart Disease, n.s. = not significant, sig. = significant.}
\label{tab:intra_model_bias_comprehensive}
\end{table}

\section{Human Evaluation}
Table \ref{tab:ttest_results} presents paired t-tests comparing human mean scores versus model-generated affective and cognitive empathy scores across demographic categories.

Table \ref{tab:african_vs_european_female} reports t-tests comparing affective and cognitive empathy scores between African female and European female groups with high school or lower education.

Table \ref{tab:human_rater_distribution} shows a detailed distribution of human ratings, including 95\% CI and rating ranges. 

\begin{table*}[!ht]
\centering
\begin{scriptsize}
\begin{tabular}{l
                r@{\,}r  r@{\,}r  r@{\,}r  r@{\,}r}
\toprule
\textbf{Category} &
\multicolumn{2}{c}{\textbf{H vs GPT (Aff)}} &
\multicolumn{2}{c}{\textbf{H vs Claude (Aff)}} &
\multicolumn{2}{c}{\textbf{H vs GPT (Cog)}} &
\multicolumn{2}{c}{\textbf{H vs Claude (Cog)}} \\
\midrule
European female  & $t=-5.25$ & $\mathbf{p=0.001}$ & $t=-0.23$ & $p=0.820$ & $t=-3.75$ & $\mathbf{p=0.005}$ & $t=-1.15$ & $p=0.280$ \\
African female   & $t=-2.59$ & $\mathbf{p=0.029}$ & $t= 0.00$ & $p=1.000$ & $t=-3.71$ & $\mathbf{p=0.005}$ & $t=-1.29$ & $p=0.229$ \\
European male    & $t=-3.00$ & $\mathbf{p=0.015}$ & $t=-0.69$ & $p=0.509$ & $t=-2.45$ & $\mathbf{p=0.037}$ & $t=-1.50$ & $p=0.168$ \\
African male     & $t=-6.00$ & $\mathbf{p<0.001}$ & $t=-0.56$ & $p=0.591$ & $t=-3.00$ & $\mathbf{p=0.015}$ & $t=-1.86$ & $p=0.096$ \\
\bottomrule
\end{tabular}
\end{scriptsize}
\caption{Paired t-test of human mean vs.\ model empathy (Affective empathy and Cognitive empathy) scores on GPT-generated responses, stratified by demographic category. Bold indicates p < 0.05.}
\label{tab:ttest_results}
\end{table*}

\begin{table*}[!ht]
\centering
\begin{tabular}{lcc}
\toprule
\textbf{Source} & \textbf{t-statistic} & \textbf{p-value} \\
\midrule
Affective Human    & -2.38 & \textbf{0.029}* \\
Affective GPT      & -1.57 & 0.138 \\
Affective Claude   & -1.90 & 0.075 \\
Cognitive Human    & -1.56 & 0.135 \\
Cognitive GPT      & -1.00 & 0.343 \\
Cognitive Claude   & -0.95 & 0.356 \\
\bottomrule
\end{tabular}
\caption{T-tests between African female and European female groups (with high school or lower education). * p < 0.05}
\label{tab:african_vs_european_female}
\end{table*}

\begin{table*}[!ht]
\centering
\begin{scriptsize}
\resizebox{\textwidth}{!}{
\begin{tabular}{lcccccccc}
\toprule
\textbf{Rater} &
\textbf{African Female} &
\textbf{African Male} &
\textbf{European Female} &
\textbf{European Male} &
\textbf{Affective Mean $\pm$ 95\% CI} &
\textbf{Affective Range} &
\textbf{Cognitive Mean $\pm$ 95\% CI} &
\textbf{Cognitive Range} \\
\midrule
Human 1 & 10 & 10 & 10 & 0  & $2.27 \pm 0.22$ & 1--3 & $2.70 \pm 0.20$ & 1--3 \\
Human 2 & 10 & 0  & 10 & 10 & $2.10 \pm 0.23$ & 1--3 & $2.27 \pm 0.29$ & 1--3 \\
Human 3 & 10 & 0  & 10 & 0  & $2.35 \pm 0.27$ & 1--3 & $2.65 \pm 0.23$ & 2--3 \\
Human 4 & 10 & 0  & 10 & 0  & $2.75 \pm 0.21$ & 2--3 & $2.25 \pm 0.30$ & 1--3 \\
\midrule
\textbf{Total} & \textbf{40} & \textbf{10} & \textbf{40} & \textbf{10} &  &  &  &  \\
\bottomrule
\end{tabular}
}
\end{scriptsize}
\caption{Detailed distribution of human ratings across demographic groups, including affective and cognitive empathy statistics (mean $\pm$ 95\% CI, based on sample standard deviation).}
\label{tab:human_rater_distribution}
\end{table*}


\clearpage
\section*{NeurIPS Paper Checklist}

\begin{enumerate}

\item {\bf Claims}
    \item[] Question: Do the main claims made in the abstract and introduction accurately reflect the paper's contributions and scope?
    \item[] Answer: \answerYes{} 
    \item[] Justification: The abstract and introduction accurately reflect the paper's contributions, which include: (1) developing an evaluation framework for assessing LLM understandability and empathy in medical diagnoses, (2) evaluating GPT-4o and Claude-3.7, (3) identifying systematic biases in both dimensions, and (4) revealing self-evaluation biases in LLM-as-a-judge approaches. The scope is clearly defined as focusing on diagnostic communication rather than clinical accuracy. 
    \item[] Guidelines:
    \begin{itemize}
        \item The answer NA means that the abstract and introduction do not include the claims made in the paper.
        \item The abstract and/or introduction should clearly state the claims made, including the contributions made in the paper and important assumptions and limitations. A No or NA answer to this question will not be perceived well by the reviewers. 
        \item The claims made should match theoretical and experimental results, and reflect how much the results can be expected to generalize to other settings. 
        \item It is fine to include aspirational goals as motivation as long as it is clear that these goals are not attained by the paper. 
    \end{itemize}

\item {\bf Limitations}
    \item[] Question: Does the paper discuss the limitations of the work performed by the authors?
    \item[] Answer: \answerYes{} 
    \item[] Justification: The paper discusses several limitations in the Conclusion section, including: narrow range of patient scenarios, small human evaluation sample size, evaluator bias issues with weak inter-rater agreement, and cultural contingency of empathy judgments. The authors also acknowledge that scenarios cannot capture full real-world variability.
    \item[] Guidelines:
    \begin{itemize}
        \item The answer NA means that the paper has no limitation while the answer No means that the paper has limitations, but those are not discussed in the paper. 
        \item The authors are encouraged to create a separate "Limitations" section in their paper.
        \item The paper should point out any strong assumptions and how robust the results are to violations of these assumptions (e.g., independence assumptions, noiseless settings, model well-specification, asymptotic approximations only holding locally). The authors should reflect on how these assumptions might be violated in practice and what the implications would be.
        \item The authors should reflect on the scope of the claims made, e.g., if the approach was only tested on a few datasets or with a few runs. In general, empirical results often depend on implicit assumptions, which should be articulated.
        \item The authors should reflect on the factors that influence the performance of the approach. For example, a facial recognition algorithm may perform poorly when image resolution is low or images are taken in low lighting. Or a speech-to-text system might not be used reliably to provide closed captions for online lectures because it fails to handle technical jargon.
        \item The authors should discuss the computational efficiency of the proposed algorithms and how they scale with dataset size.
        \item If applicable, the authors should discuss possible limitations of their approach to address problems of privacy and fairness.
        \item While the authors might fear that complete honesty about limitations might be used by reviewers as grounds for rejection, a worse outcome might be that reviewers discover limitations that aren't acknowledged in the paper. The authors should use their best judgment and recognize that individual actions in favor of transparency play an important role in developing norms that preserve the integrity of the community. Reviewers will be specifically instructed to not penalize honesty concerning limitations.
    \end{itemize}

\item {\bf Theory assumptions and proofs}
    \item[] Question: For each theoretical result, does the paper provide the full set of assumptions and a complete (and correct) proof?
    \item[] Answer: \answerNA{} 
    \item[] Justification: This paper does not include theoretical results requiring formal proofs. It is an empirical evaluation study using established readability metrics and empathy assessment frameworks.
    \item[] Guidelines:
    \begin{itemize}
        \item The answer NA means that the paper does not include theoretical results. 
        \item All the theorems, formulas, and proofs in the paper should be numbered and cross-referenced.
        \item All assumptions should be clearly stated or referenced in the statement of any theorems.
        \item The proofs can either appear in the main paper or the supplemental material, but if they appear in the supplemental material, the authors are encouraged to provide a short proof sketch to provide intuition. 
        \item Inversely, any informal proof provided in the core of the paper should be complemented by formal proofs provided in appendix or supplemental material.
        \item Theorems and Lemmas that the proof relies upon should be properly referenced. 
    \end{itemize}

    \item {\bf Experimental result reproducibility}
    \item[] Question: Does the paper fully disclose all the information needed to reproduce the main experimental results of the paper to the extent that it affects the main claims and/or conclusions of the paper (regardless of whether the code and data are provided or not)?
    \item[] Answer: \answerYes{} 
    \item[] Justification: The paper provides detailed methodology including: prompt templates (Section 3), specific readability metrics with formulas (Table 1), empathy evaluation prompts (code listings), demographic combinations (156 prompts), and statistical analysis methods. The framework and evaluation pipeline are thoroughly documented.
    \item[] Guidelines:
    \begin{itemize}
        \item The answer NA means that the paper does not include experiments.
        \item If the paper includes experiments, a No answer to this question will not be perceived well by the reviewers: Making the paper reproducible is important, regardless of whether the code and data are provided or not.
        \item If the contribution is a dataset and/or model, the authors should describe the steps taken to make their results reproducible or verifiable. 
        \item Depending on the contribution, reproducibility can be accomplished in various ways. For example, if the contribution is a novel architecture, describing the architecture fully might suffice, or if the contribution is a specific model and empirical evaluation, it may be necessary to either make it possible for others to replicate the model with the same dataset, or provide access to the model. In general. releasing code and data is often one good way to accomplish this, but reproducibility can also be provided via detailed instructions for how to replicate the results, access to a hosted model (e.g., in the case of a large language model), releasing of a model checkpoint, or other means that are appropriate to the research performed.
        \item While NeurIPS does not require releasing code, the conference does require all submissions to provide some reasonable avenue for reproducibility, which may depend on the nature of the contribution. For example
        \begin{enumerate}
            \item If the contribution is primarily a new algorithm, the paper should make it clear how to reproduce that algorithm.
            \item If the contribution is primarily a new model architecture, the paper should describe the architecture clearly and fully.
            \item If the contribution is a new model (e.g., a large language model), then there should either be a way to access this model for reproducing the results or a way to reproduce the model (e.g., with an open-source dataset or instructions for how to construct the dataset).
            \item We recognize that reproducibility may be tricky in some cases, in which case authors are welcome to describe the particular way they provide for reproducibility. In the case of closed-source models, it may be that access to the model is limited in some way (e.g., to registered users), but it should be possible for other researchers to have some path to reproducing or verifying the results.
        \end{enumerate}
    \end{itemize}

\item {\bf Open access to data and code}
    \item[] Question: Does the paper provide open access to the data and code, with sufficient instructions to faithfully reproduce the main experimental results, as described in supplemental material?
    \item[] Answer: \answerYes{} 
    \item[] Justification: uploaded in the supplementary material
    \item[] Guidelines:
    \begin{itemize}
        \item The answer NA means that paper does not include experiments requiring code.
        \item Please see the NeurIPS code and data submission guidelines (\url{https://nips.cc/public/guides/CodeSubmissionPolicy}) for more details.
        \item While we encourage the release of code and data, we understand that this might not be possible, so “No” is an acceptable answer. Papers cannot be rejected simply for not including code, unless this is central to the contribution (e.g., for a new open-source benchmark).
        \item The instructions should contain the exact command and environment needed to run to reproduce the results. See the NeurIPS code and data submission guidelines (\url{https://nips.cc/public/guides/CodeSubmissionPolicy}) for more details.
        \item The authors should provide instructions on data access and preparation, including how to access the raw data, preprocessed data, intermediate data, and generated data, etc.
        \item The authors should provide scripts to reproduce all experimental results for the new proposed method and baselines. If only a subset of experiments are reproducible, they should state which ones are omitted from the script and why.
        \item At submission time, to preserve anonymity, the authors should release anonymized versions (if applicable).
        \item Providing as much information as possible in supplemental material (appended to the paper) is recommended, but including URLs to data and code is permitted.
    \end{itemize}

\item {\bf Experimental setting/details}
    \item[] Question: Does the paper specify all the training and test details (e.g., data splits, hyperparameters, how they were chosen, type of optimizer, etc.) necessary to understand the results?
    \item[] Answer: \answerYes{} 
    \item[] Justification: The paper reports appropriate statistical measures including error bars (±1 SD), p-values, effect sizes (Cohen's d), ANOVA results, and t-test statistics. Comprehensive statistical validation is provided in the appendix with detailed significance testing results.
    \item[] Guidelines:
    \begin{itemize}
        \item The answer NA means that the paper does not include experiments.
        \item The experimental setting should be presented in the core of the paper to a level of detail that is necessary to appreciate the results and make sense of them.
        \item The full details can be provided either with the code, in appendix, or as supplemental material.
    \end{itemize}

\item {\bf Experiment statistical significance}
    \item[] Question: Does the paper report error bars suitably and correctly defined or other appropriate information about the statistical significance of the experiments?
    \item[] Answer: \answerYes{} 
    \item[] Justification: The paper reports appropriate statistical measures including error bars (±1 SD), p-values, effect sizes (Cohen's d), ANOVA results, and t-test statistics. Comprehensive statistical validation is provided in the appendix with detailed significance testing results.
    \item[] Guidelines:
    \begin{itemize}
        \item The answer NA means that the paper does not include experiments.
        \item The authors should answer "Yes" if the results are accompanied by error bars, confidence intervals, or statistical significance tests, at least for the experiments that support the main claims of the paper.
        \item The factors of variability that the error bars are capturing should be clearly stated (for example, train/test split, initialization, random drawing of some parameter, or overall run with given experimental conditions).
        \item The method for calculating the error bars should be explained (closed form formula, call to a library function, bootstrap, etc.)
        \item The assumptions made should be given (e.g., Normally distributed errors).
        \item It should be clear whether the error bar is the standard deviation or the standard error of the mean.
        \item It is OK to report 1-sigma error bars, but one should state it. The authors should preferably report a 2-sigma error bar than state that they have a 96\% CI, if the hypothesis of Normality of errors is not verified.
        \item For asymmetric distributions, the authors should be careful not to show in tables or figures symmetric error bars that would yield results that are out of range (e.g. negative error rates).
        \item If error bars are reported in tables or plots, The authors should explain in the text how they were calculated and reference the corresponding figures or tables in the text.
    \end{itemize}

\item {\bf Experiments compute resources}
    \item[] Question: For each experiment, does the paper provide sufficient information on the computer resources (type of compute workers, memory, time of execution) needed to reproduce the experiments?
    \item[] Answer: \answerYes{} 
    \item[] Justification: The paper mentioned the usage of API calls to the corresponding LLM models.
    \item[] Guidelines:
    \begin{itemize}
        \item The answer NA means that the paper does not include experiments.
        \item The paper should indicate the type of compute workers CPU or GPU, internal cluster, or cloud provider, including relevant memory and storage.
        \item The paper should provide the amount of compute required for each of the individual experimental runs as well as estimate the total compute. 
        \item The paper should disclose whether the full research project required more compute than the experiments reported in the paper (e.g., preliminary or failed experiments that didn't make it into the paper). 
    \end{itemize}
    
\item {\bf Code of ethics}
    \item[] Question: Does the research conducted in the paper conform, in every respect, with the NeurIPS Code of Ethics \url{https://neurips.cc/public/EthicsGuidelines}?
    \item[] Answer: \answerYes{}
    \item[] Justification: We ensure that our experiments comply with the NeurIPS Code of Ethics in all respects.
    \item[] Guidelines:
    \begin{itemize}
        \item The answer NA means that the authors have not reviewed the NeurIPS Code of Ethics.
        \item If the authors answer No, they should explain the special circumstances that require a deviation from the Code of Ethics.
        \item The authors should make sure to preserve anonymity (e.g., if there is a special consideration due to laws or regulations in their jurisdiction).
    \end{itemize}

\item {\bf Broader impacts}
    \item[] Question: Does the paper discuss both potential positive societal impacts and negative societal impacts of the work performed?
    \item[] Answer: \answerYes{} 
    \item[] Justification: The paper includes a dedicated ``Broader Impacts" section discussing both positive impacts (improving healthcare access, supporting clinicians) and negative impacts (amplifying health inequities, undermining trust among vulnerable groups). It emphasizes the need for transparent, bias-aware evaluation before deployment
    \item[] Guidelines:
    \begin{itemize}
        \item The answer NA means that there is no societal impact of the work performed.
        \item If the authors answer NA or No, they should explain why their work has no societal impact or why the paper does not address societal impact.
        \item Examples of negative societal impacts include potential malicious or unintended uses (e.g., disinformation, generating fake profiles, surveillance), fairness considerations (e.g., deployment of technologies that could make decisions that unfairly impact specific groups), privacy considerations, and security considerations.
        \item The conference expects that many papers will be foundational research and not tied to particular applications, let alone deployments. However, if there is a direct path to any negative applications, the authors should point it out. For example, it is legitimate to point out that an improvement in the quality of generative models could be used to generate defakes for disinformation. On the other hand, it is not needed to point out that a generic algorithm for optimizing neural networks could enable people to train models that generate Deepfakes faster.
        \item The authors should consider possible harms that could arise when the technology is being used as intended and functioning correctly, harms that could arise when the technology is being used as intended but gives incorrect results, and harms following from (intentional or unintentional) misuse of the technology.
        \item If there are negative societal impacts, the authors could also discuss possible mitigation strategies (e.g., gated release of models, providing defenses in addition to attacks, mechanisms for monitoring misuse, mechanisms to monitor how a system learns from feedback over time, improving the efficiency and accessibility of ML).
    \end{itemize}
    
\item {\bf Safeguards}
    \item[] Question: Does the paper describe safeguards that have been put in place for responsible release of data or models that have a high risk for misuse (e.g., pretrained language models, image generators, or scraped datasets)?
    \item[] Answer: \answerNA{} 
    \item[] Justification: The evaluation framework itself poses no risk and could help implement safeguards in medical AI systems.
    \item[] Guidelines:
    \begin{itemize}
        \item The answer NA means that the paper poses no such risks.
        \item Released models that have a high risk for misuse or dual-use should be released with necessary safeguards to allow for controlled use of the model, for example by requiring that users adhere to usage guidelines or restrictions to access the model or implementing safety filters. 
        \item Datasets that have been scraped from the Internet could pose safety risks. The authors should describe how they avoided releasing unsafe images.
        \item We recognize that providing effective safeguards is challenging, and many papers do not require this, but we encourage authors to take this into account and make a best faith effort.
    \end{itemize}

\item {\bf Licenses for existing assets}
    \item[] Question: Are the creators or original owners of assets (e.g., code, data, models), used in the paper, properly credited and are the license and terms of use explicitly mentioned and properly respected?
    \item[] Answer: \answerYes{} 
    \item[] Justification: 
    The paper uses established commercial LLM APIs (properly cited), and the EmotionQueen framework, whose repository is provided in Appendix A as a footnote, and all assets are used in accordance with their terms of use. 
    \item[] Guidelines:
    \begin{itemize}
        \item The answer NA means that the paper does not use existing assets.
        \item The authors should cite the original paper that produced the code package or dataset.
        \item The authors should state which version of the asset is used and, if possible, include a URL.
        \item The name of the license (e.g., CC-BY 4.0) should be included for each asset.
        \item For scraped data from a particular source (e.g., website), the copyright and terms of service of that source should be provided.
        \item If assets are released, the license, copyright information, and terms of use in the package should be provided. For popular datasets, \url{paperswithcode.com/datasets} has curated licenses for some datasets. Their licensing guide can help determine the license of a dataset.
        \item For existing datasets that are re-packaged, both the original license and the license of the derived asset (if it has changed) should be provided.
        \item If this information is not available online, the authors are encouraged to reach out to the asset's creators.
    \end{itemize}

\item {\bf New assets}
    \item[] Question: Are new assets introduced in the paper well documented and is the documentation provided alongside the assets?
    \item[] Answer: \answerNA{} 
    \item[] Justification: 
     The paper does not release new assets. 
    \item[] Guidelines:
    \begin{itemize}
        \item The answer NA means that the paper does not release new assets.
        \item Researchers should communicate the details of the dataset/code/model as part of their submissions via structured templates. This includes details about training, license, limitations, etc. 
        \item The paper should discuss whether and how consent was obtained from people whose asset is used.
        \item At submission time, remember to anonymize your assets (if applicable). You can either create an anonymized URL or include an anonymized zip file.
    \end{itemize}

\item {\bf Crowdsourcing and research with human subjects}
    \item[] Question: For crowdsourcing experiments and research with human subjects, does the paper include the full text of instructions given to participants and screenshots, if applicable, as well as details about compensation (if any)? 
    \item[] Answer: \answerYes{} 
    \item[] Justification: The paper includes human evaluation with four annotators from the research team. 
    \item[] Guidelines:
    \begin{itemize}
        \item The answer NA means that the paper does not involve crowdsourcing nor research with human subjects.
        \item Including this information in the supplemental material is fine, but if the main contribution of the paper involves human subjects, then as much detail as possible should be included in the main paper. 
        \item According to the NeurIPS Code of Ethics, workers involved in data collection, curation, or other labor should be paid at least the minimum wage in the country of the data collector. 
    \end{itemize}

\item {\bf Institutional review board (IRB) approvals or equivalent for research with human subjects}
    \item[] Question: Does the paper describe potential risks incurred by study participants, whether such risks were disclosed to the subjects, and whether Institutional Review Board (IRB) approvals (or an equivalent approval/review based on the requirements of your country or institution) were obtained?
    \item[] Answer: \answerNA{}{} 
    \item[] Justification: no external human participant, only the authors have participated in this project.
    \item[] Guidelines:
    \begin{itemize}
        \item The answer NA means that the paper does not involve crowdsourcing nor research with human subjects.
        \item Depending on the country in which research is conducted, IRB approval (or equivalent) may be required for any human subjects research. If you obtained IRB approval, you should clearly state this in the paper. 
        \item We recognize that the procedures for this may vary significantly between institutions and locations, and we expect authors to adhere to the NeurIPS Code of Ethics and the guidelines for their institution. 
        \item For initial submissions, do not include any information that would break anonymity (if applicable), such as the institution conducting the review.
    \end{itemize}

\item {\bf Declaration of LLM usage}
    \item[] Question: Does the paper describe the usage of LLMs if it is an important, original, or non-standard component of the core methods in this research? Note that if the LLM is used only for writing, editing, or formatting purposes and does not impact the core methodology, scientific rigorousness, or originality of the research, declaration is not required.
    \item[] Answer: \answerYes{} 
    \item[] Justification: LLMs (GPT-4o and Claude-3.7) are the core focus of this research - both as the systems being evaluated and as judges in the LLM-as-a-judge evaluation framework. Their usage is thoroughly documented and central to the methodology.
    \item[] Guidelines:
    \begin{itemize}
        \item The answer NA means that the core method development in this research does not involve LLMs as any important, original, or non-standard components.
        \item Please refer to our LLM policy (\url{https://neurips.cc/Conferences/2025/LLM}) for what should or should not be described.
    \end{itemize}

\end{enumerate}

\end{document}